\documentclass{article} 
\usepackage{iclr2024_conference,times}

\usepackage{amsmath,amsfonts,bm}

\def\eqref#1{equation~\ref{#1}}

\def\1{\bm{1}}

\DeclareMathAlphabet{\mathsfit}{\encodingdefault}{\sfdefault}{m}{sl}
\SetMathAlphabet{\mathsfit}{bold}{\encodingdefault}{\sfdefault}{bx}{n}

\usepackage{hyperref}
\usepackage{url}
\usepackage{fancyhdr} 

\fancypagestyle{plain}{
    \fancyhf{} 
    \fancyhead[L]{\textit{Paper under review}} 
    \renewcommand{\headrulewidth}{0.4pt} 
}
\title{Context-aware Prompt Tuning: Advancing In-Context Learning with Adversarial Methods}
\iclrfinalcopy

\author{
    Tsachi Blau\textsuperscript{\dag}\thanks{\texttt{tsachiblau@campus.technion.ac.il}} \quad 
    Moshe Kimhi\textsuperscript{\dag} \quad 
    Yonatan Belinkov\textsuperscript{\dag} \quad 
    Alexander Bronstein\textsuperscript{\dag} \quad 
    Chaim Baskin\textsuperscript{\ddag}
}

\usepackage{graphicx}
\usepackage{multirow}
\usepackage{multicol}
\usepackage{dblfloatfix}
\usepackage{algpseudocode}
\usepackage{algorithm}
\usepackage{enumitem}
\usepackage{lipsum} 
\usepackage{wrapfig} 
\usepackage{xcolor}         
\usepackage{color, colortbl}
\usepackage{makecell}
\usepackage{pifont}
\usepackage{amsmath}
\usepackage{cleveref}
\usepackage{tabularx}
\usepackage{pifont}
\usepackage{tabularx}

\newcommand{\AlgoName}{CPT }
\newcommand{\AlgoNameNoSpace}{CPT}
\newcommand{\bloom}{BLOOM }
\newcommand{\gpt}{GPT-j }
\newcommand{\llama}{Llama3 }
\newcommand{\bloomNoSpace}{BLOOM}
\newcommand{\gptNoSpace}{GPT-j}
\newcommand{\llamaNoSpace}{Llama3}

\newcommand{\sst}{SST-2 }
\newcommand{\sstNoSpace}{SST-2}
\newcommand{\agnews}{AG News }
\newcommand{\agnewsNoSpace}{AG News}
\newcommand{\dbpedia}{DBpedia }
\newcommand{\dbpediaNoSpace}{DBpedia}
\newcommand{\trec}{TREC }
\newcommand{\trecNoSpace}{TREC}
\newcommand{\cmark}{\ding{51}}%
\newcommand{\xmark}{\ding{55}}%

\definecolor{gray}{RGB}{0, 0, 0}
\definecolor{golden}{RGB}{255, 215, 0}

\newcommand{\upFigureAboveFig}{-1.2 cm}
\newcommand{\upFigureBelowFig}{-0.7 cm}
\newcommand{\upFigureBelowCap}{-0.2 cm}

\newcommand{\downFigureAboveFig}{-0.0 cm}
\newcommand{\downFigureBelowFig}{\upFigureBelowFig}
\newcommand{\downFigureBelowCap}{-0.2 cm}

\newcommand{\upTabAboveCap}{-1.4 cm}
\newcommand{\upTabBelowCap}{-0.0 cm}
\newcommand{\upTabBelowTab}{-0.0 cm}

\begin{document}

\maketitle
\renewcommand{\thefootnote}{\fnsymbol{footnote}}
\footnotetext[2]{Department of Computer Science Technion - Israel Institute of Technology}
\footnotetext[3]{School of Electrical and Computer Engineering - Ben-Gurion University of the Negev}

\begin{abstract}

Fine-tuning Large Language Models (LLMs) typically involves updating at least a few billions of parameters.
A more parameter-efficient approach is Prompt Tuning (PT), which updates only a few learnable tokens, and differently, In-Context Learning (ICL) adapts the model to a new task by simply including examples in the input without any training.
When applying optimization-based methods, such as fine-tuning and PT for few-shot learning, the model is specifically adapted to the small set of training examples, whereas ICL leaves the model unchanged. 
This distinction makes traditional learning methods more prone to overfitting; in contrast, ICL is less sensitive to the few-shot scenario.
While ICL is not prone to overfitting, it does not fully extract the information that exists in the training examples.
This work introduces Context-aware Prompt Tuning (CPT), a method inspired by ICL, PT, and adversarial attacks. 
We build on the ICL strategy of concatenating examples before the input, but we extend this by PT-like learning, refining the context embedding through iterative optimization to extract deeper insights from the training examples.
We carefully modify specific context tokens, considering the unique structure of input and output formats. 
Inspired by adversarial attacks, we adjust the input based on the labels present in the context, focusing on minimizing, rather than maximizing, the loss.
Moreover, we apply a projected gradient descent algorithm to keep token embeddings close to their original values, under the assumption that the user-provided data is inherently valuable. 
Our method has been shown to achieve superior accuracy across multiple classification tasks using various LLM models.

\end{abstract}

\begin{figure}[hb!]
    \centering
    \includegraphics[width=\textwidth]{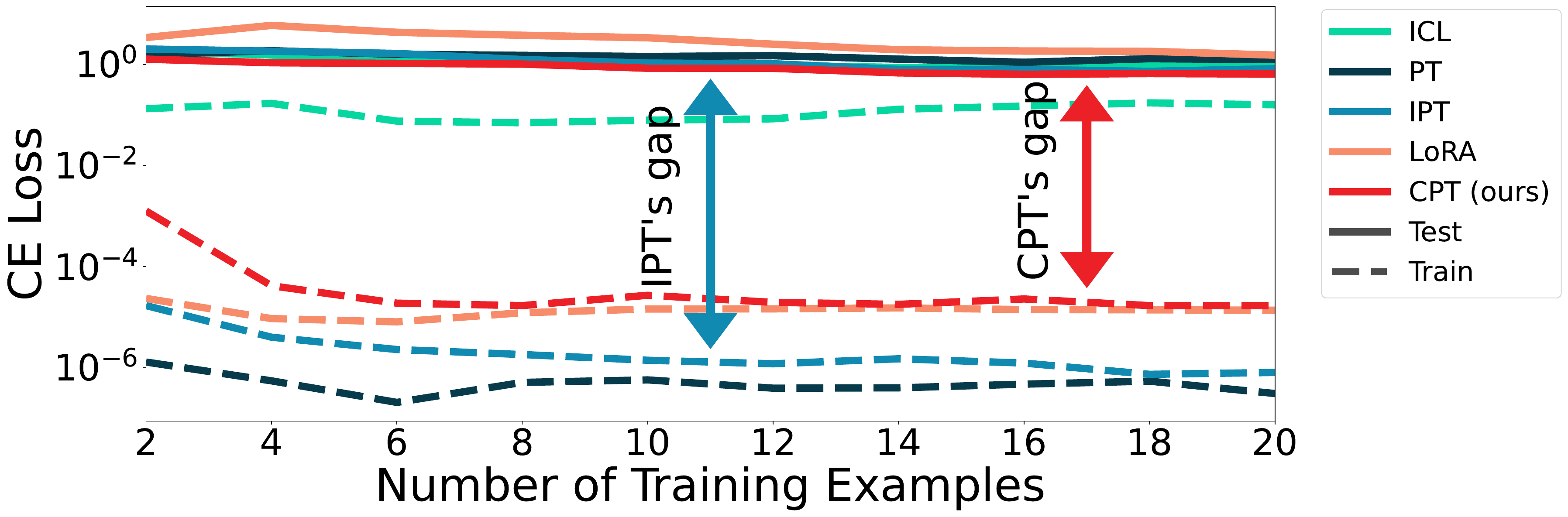}
    \caption{\textbf{Overfitting Comparison: CPT vs. Baselines} Visualizing the train-test loss gap across various methods and training set sizes using the \gpt model on the \dbpedia dataset. For each model, there are two loss graphs: one for train loss (dotted line) and one for test loss (solid line). \AlgoName performs better in mitigating overfitting compared to optimization-based methods. Despite a relatively higher training loss, \AlgoName achieves the lowest test loss.}
    \label{fig:eval_loss}
\end{figure}

\begin{figure}[t!]
    \centering
    \vspace{\upFigureAboveFig}
    \includegraphics[width=\textwidth]{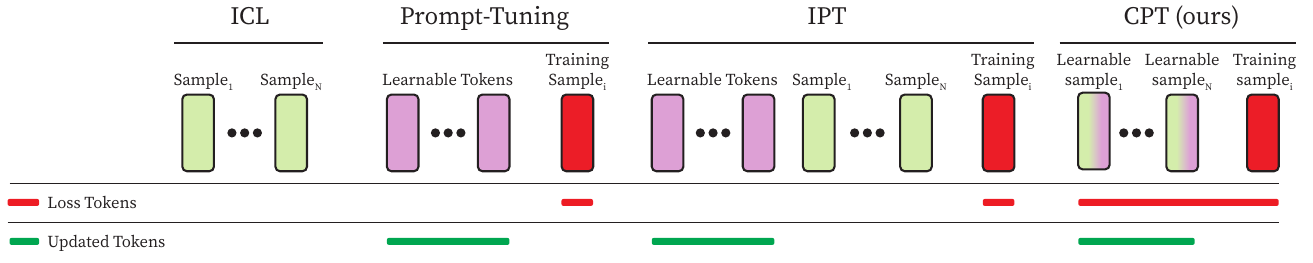}
    \vspace{\upFigureBelowFig}
    \caption{\textbf{Comparison of Baseline Algorithms and Token Utilization.} We highlight the key differences between \AlgoName and the baselines, focusing on ICL, PT, and IPT. 
    For each method, we emphasize two types of tokens: those used for loss calculation (red line) and those updated during optimization (green line). 
    \AlgoName features $\emph{`Learnable Sample}_i'$ in dual colors, reflecting their progression from $`\emph{Sample tokens}'$ to $`\emph{Learnable tokens}'$ as they are optimized.
    }
    \vspace{\upFigureBelowCap}
    \label{fig:baselines_comparison}
\end{figure}

\section{Introduction}
\label{sec:introduction}


Fine-tuning Large Language Models (LLMs) is a widely used technique that adapts models to specific tasks by modifying all their parameters. 
Despite its effectiveness, this approach requires handling billions of parameters, which can be prohibitively expensive and inefficient, especially in terms of computational resources and storage, making it challenging to scale effectively.

To address the limitations of fine-tuning, several parameter-efficient methods have been introduced. 
Low-Rank Adaptation (LoRA) \citep{hu2021lora} reduces the number of trainable parameters by learning a low-rank decomposition. 
However, it still requires a portion of the model's weights, which remains burdensome, particularly since state-of-the-art models like \llama \citep{llama3modelcard} and GPT-4 \citep{openai2024chatgpt} typically range from 7 billion to 1.7 trillion parameters. 
Another approach, Prompt Tuning (PT) \citep{lester2021power}, offers a more efficient solution by updating a small set of learnable token embeddings, which are concatenated before the input, while leaving the LLM’s weights completely untouched.
Alternatively, In-Context Learning (ICL) \citep{brown2020language} adjusts the model to new tasks without any parameter updates, relying on the straightforward concatenation of training examples with the input context.
Despite its computational efficiency, recent studies \cite{zhang2022opt, sun2023does,perez2021true} indicate that ICL falls short compared to supervised fine-tuning methods. 
To leverage the strengths of both PT and ICL, Instruction Prompt Tuning (IPT) \citep{singhal2022large} was introduced. This approach involves concatenating both learnable tokens and context to the input, training only the learnable tokens while keeping the context and model weights frozen.

Despite the recent advancements in parameter-efficient methods, determining the optimal method for few-shot learning remains highly unsettled. 
On one hand, optimization-based methods such as fine-tuning, LoRA, PT, and IPT are prone to overfitting, especially in few-shot settings where the number of trainable parameters is large -- a condition known to exacerbate overfitting, as demonstrated in \cref{fig:eval_loss}. 
Meanwhile, In-Context Learning (ICL) mitigates overfitting by avoiding model parameter updates; however, it does not match the performance of other methods.
Consequently, the most effective method for various different scenarios remains uncertain \citep{sun2023does}.

Context-aware Prompt Tuning (CPT), fuses concepts from In-Context Learning (ICL), Prompt Tuning (PT), and adversarial attacks \citep{blau2022threat, blau2023classifier, carlini2017adversarial, athalye2018synthesizing,biggio2013evasion,goodfellow2014explaining,kurakin2016adversarial,nguyen2015deep,madry2017towards,rebuffi2021fixing,gowal2020uncovering} into a cohesive approach, with the main differences from the baselines illustrated \cref{fig:baselines_comparison}.
CPT follows the ICL technique of concatenating training examples prior to the input. 
Similarly to PT, CPT updates only the context token embeddings through iterative optimization, leveraging again the training examples present in the context.
However, CPT carefully refines the context tokens while accounting for the context's unique structure, keeping the label tokens intact, preserving their role as the ground truth.
To effectively reduce overfitting and enhance performance, CPT adopts two strategies inspired by adversarial attacks: incorporating context labels into the loss function and applying projection after updating the token embeddings. 
By including context labels in the loss, CPT refines input adjustments, guiding the model to optimize the entire context rather than focusing solely on the training label. 
To further mitigate overfitting, projected gradient descent is applied after each optimization step. 
This method ensures that token embedding updates remain within a controlled range, preserving proximity to their original values, under the assumption that user-provided examples are valuable. 
Additionally, CPT employs a loss weighting approach leverages recency bias -- a phenomenon highlighted by \cite{zhao2021calibrate}, where the model tends to prioritize examples located nearer the end of the context.
We recommend leveraging this property by applying an exponentially decaying weight to examples as they approach the beginning of the context, thereby increasing the emphasis on more recent examples in the optimization process.

\begin{figure}[t!]
    \centering
    \vspace{\upFigureAboveFig}
    \includegraphics[width=\textwidth]{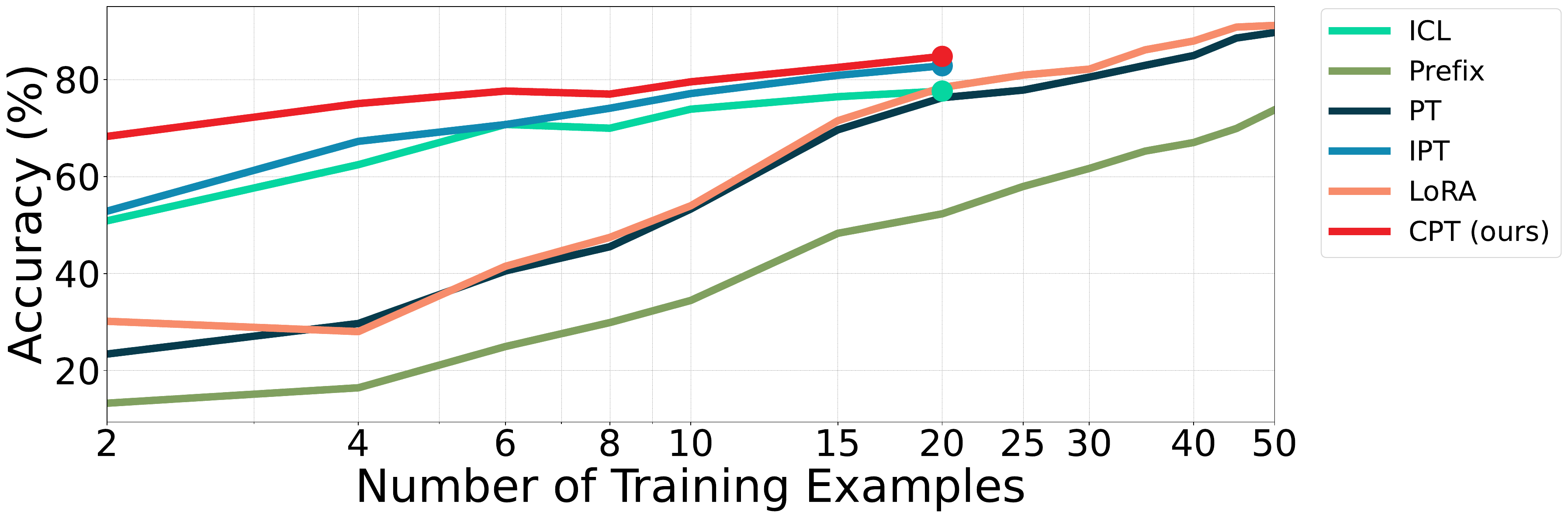}
    \vspace{\upFigureBelowFig}
    
    \caption{\textbf{Few-Shot Methods Comparison.} We compare \AlgoName using the \gpt model and the \dbpedia dataset to baselines in few-shot settings, showing that it particularly excels when dealing with a limited number of examples. Additionally, we show that context-based methods hit memory constraints (marked with a dot) as the number of training examples rises beyond a certain level.}
    
    \vspace{\upFigureBelowCap}
    
    \label{fig:few_shots}
\end{figure}

We validate our method through a comprehensive evaluation of several classification tasks and include extensive ablations.
We demonstrate that CPT outperforms other baselines across nearly every scenario, as shown in \cref{fig:few_shots}. 
We use diverse templates and seeds, which is crucial due to ICL's sensitivity to training examples and format selection, as highlighted by \cite{sun2023does, zhao2021calibrate}. 
Additionally, we assess our method on a novel classification task, termed 'Set Classification' where it shows robust performance and effectiveness on tasks the model has not previously encountered.

To summarize, our key contributions are as follows:
\begin{itemize}
    \item We propose a novel few-shot method called Context-aware Prompt Tuning that enhances ICL with PT and adversarial methodologies. 
    Our method carefully optimizes the context tokens while accounting for their unique structure.
    \item Our method incorporates ground truth labels from the context into the loss term, optimizes with projected gradient descent, and applies recency-bias-inspired loss weighting.
    \item We introduce a new classification task termed `Set Classification', demonstrating our method's effectiveness on tasks the model has not previously encountered.
    \item We achieve state-of-the-art results on several classification datasets and provide extensive ablation studies for each design choice of our method.
\end{itemize}

\section{Related Work}
\label{sec:related_work}

\textbf{Fine-Tuning} Fine-tuning is a popular and effective method for adjusting LLMs to specific tasks. Standard fine-tuning \citep{radford2019language, brown2020language, howard2018universal, liu2019roberta, lan2019albert, raffel2020exploring, sun2019fine} retrains the model with new data. However, a key disadvantage is the large number of parameters that must be stored.

\textbf{Efficient Fine-Tuning} To alleviate the computational burden of fine-tuning, Adapter-BERT \citep{houlsby2019parameter} proposes training only the adapter layers inserted into the model, while BitFit \citep{zaken2021bitfit} focuses on fine-tuning just the bias terms. Delta Tuning \citep{ding2022delta} explores parameter-efficient methods that adjust only a small portion of a model's parameters. Low-Rank Adaptation methods (LoRA) \citep{hu2021lora} introduces a novel low-rank adaptation technique, where additional low-rank matrices are added to the weights during training. This allows the model to fine-tune only these matrices, reducing the number of trainable parameters significantly. VERA \citep{kopiczko2023vera} builds on LoRA by incorporating adaptive learning rates. Compacter \cite{karimi2021compacter} leverages hypercomplex layers, and LoRA-Pro \citep{wang2024lora} further refines optimization. Despite these advancements, large models like GPT-3, which contain $175B$ parameters, require updating millions of parameters, such as 17.5M for LoRA.

\textbf{Prompt Tuning (PT)} Unlike fine-tuning methods, PT reduces the number of trainable parameters by introducing learnable tokens optimized while keeping the model's weights frozen. \cite{lester2021power} propose appending continuous prompts to the input and optimizing them, while P-tuning \citep{liu2023gpt} and Prefix Tuning \citep{li2021prefix} extend this concept by incorporating learnable tokens at intermediate layers. More recently, \cite{wang2023multitask} introduced the idea of training a single prompt to be shared across multiple tasks. Although these methods significantly reduce the number of trainable parameters, they face challenges in few-shot learning \cite{gu2021ppt} and provide limited interpretability for the learned continuous tokens \citep{ghosal2024intcoop, khashabi2021prompt, deng2022rlprompt}.

\textbf{In-Context Learning (ICL)} In contrast to earlier methods, ICL \citep{brown2020language} avoids optimization entirely. Instead, it concatenates task-specific examples before the input, allowing the model to learn a new task purely through observation, leveraging its pre-trained knowledge. Despite its advantages, ICL has limitations, often underperforming compared to optimization-based methods \citep{liu2022few, peng2023does, sun2023does}.

\textbf{Instruction Prompt Tuning (IPT)} 
IPT \citep{singhal2022large} combines key elements of PT and ICL, utilizing learnable tokens that are optimized during training alongside static context tokens, similar to ICL.
The concept of using both soft and hard prompts was previously introduced by PPT \citep{gu2021ppt} and PTR \citep{han2022ptr}. 
Yet, IPT has struggled to consistently surpass PT in performance \citep{sun2023does}.
While our method shares similarities with IPT, we focus on optimizing context tokens without introducing additional learnable tokens, and we are also leveraging context labels in the process. 
Another key difference lies in the optimization process, where our loss includes a regularization term, and we employ projected gradient descent to ensure the output stays close to the user-supplied reliable input.

\section{Our Method}
\label{sec:our_method}

\subsection{Input Preparation}
\label{sec:our_method:input}
\begin{figure}[t!]
    \centering
    \vspace{\upFigureAboveFig}
    
    \includegraphics[width=\textwidth]{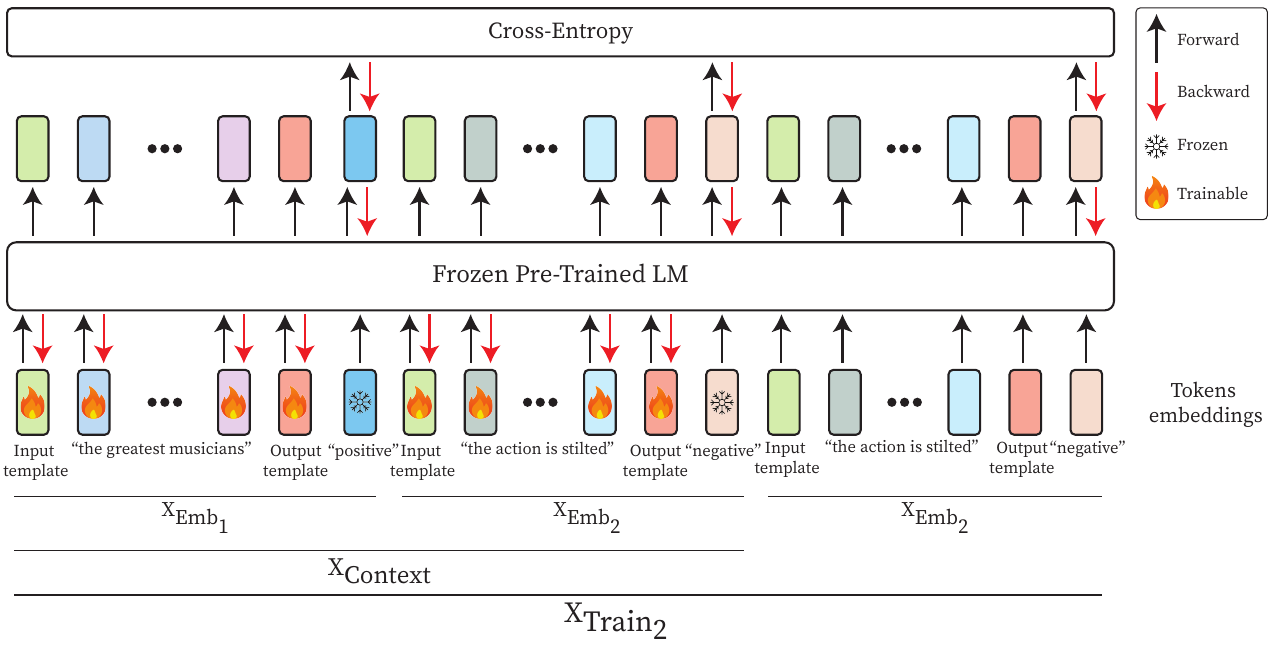}
    \vspace{\upFigureBelowFig}
    \caption{\textbf{Overview of \AlgoName Training Process.} We begin by arranging the data. We concatenate all of the training examples that were embedded into the input-output templates $[X_{\text{Emb}_i}]_{i=1}^{N}$, creating $X_{\text{Context}}$. To this, we append a randomly selected training example, in this case $X_{\text{Emb}_2}$, to form the complete training example $X_{\text{Train}_2}$. For the training process, the input is passed through the frozen LLM, and the loss is calculated using all labels present in $X_{\text{Train}_2}$, covering both the context and training labels. The context is updated, but its labels remain unchanged.
    }
    \vspace{\upFigureBelowCap}
    
    \label{fig:our_training}
\end{figure}

Our method takes as input a few-shot classification dataset containing $N$ examples. Each example consists of a pairing of $x$ (an instruction) and $y$ (a label). We embed $(x, y)$ using input, output, and separation templates, converting them into readable text that LLMs better understand, as done in ICL \cite{brown2020language}. The input and output templates, denoted $T_i$ and $T_o$, along with separators $S_{\text{intra}}$ and $S_{\text{inter}}$, are provided in appendix F.
To embed a single example ($x, y$) using the template, we concatenate the input $x$ embedded in $T_i$ with $S_{\text{intra}}$, followed by the output $y$ embedded in $T_o$, and finally $S_{\text{inter}}$, resulting in $X_{\text{Emb}_i}=[T_i(x_i), S_{\text{intra}}, T_o(y_i), S_{\text{inter}}]$. 
To generate the complete context, we concatenate all $X_{\text{Emb}_i}$, forming 
$X_{\text{Context}} = [X_{\text{Emb}_i} ]_{i=1}^{N}$.
To construct a complete training example, we randomly select an embedded example from the training set $X_{\text{Emb}_i}$, and concatenate it after the context, resulting $X_{\text{Train}_i} = [X_{\text{Context}}, X_{\text{Emb}_i}]$, which is then fed into the LLM.
This process is also visualized in \cref{fig:our_training}.

Above, we described how we construct a training example $X_{\text{Train}_i}$, as a text sequence. 
However, before feeding it into the model, we must process the text through a tokenizer, which splits the text into tokens and returns an embedding vector for each token. 
Each example contains six types of tokens: input, input template, intra-separator, output, output template, and inter-separator. 
For simplicity, we ignore the separators and the fact that each part usually contains multiple tokens.
For each training example $i$ and its sub-example $k$, we focus on four token types: $t_{\text{I}_{i}}^{(k)}, t_{\text{IT}_{i}}^{(k)}, t_{\text{O}_{i}}^{(k)}, t_{\text{OT}_{i}}^{(k)}$, which represent the input, input template, output, and output template, respectively.
Each training example $i$ consists of $N+1$ sub-examples, $N$ sub-examples in the context and one training sub-example at the end.

\subsection{Optimization}
\label{sec:our_method:optimization}
In this section, we discuss the optimization process of our method, which is inspired by Adversarial Attacks (AT) \cite{madry2017towards}. 
Each AT step can be divided into two parts: optimization and restriction. 
In the first step, an attacker modifies an image to cause incorrect classification, where in the second step, the attack limits its changes to avoid detection. 
Similarly, our method is split into two parts: optimization, including loss design, as discussed in \cref{sec:our_method:loss}, and controlling the token updates, as detailed in \cref{sec:our_method:updates}.

\subsubsection{Loss Design}
\label{sec:our_method:loss}

The optimization process modifies the input embedding to help the classification.
To achieve this, we introduce a new loss for each training example $X_{\text{Train}_i}$. This loss incorporates all the context sub-example labels $X_{\text{Context}}$, or more formally, we use $t_{\text{O}_{i}}^{(k)}$ for all $k \in [1, N]$.
We use these tokens and the model's predictions for those tokens, $\hat{t}_{\text{O}_{i}}^{(k)}$, as shown in \cref{loss_context}.

\begin{equation}
\begin{aligned}
L_{\text{Context}_i} = \sum_{k=1}^{N} & \omega_k \cdot \text{CrossEntropy}(\hat{t}_{\text{O}_{i}}^{(k)}, t_{\text{O}_{i}}^{(k)}) \\
\end{aligned}
\label{loss_context}
\end{equation}

In addition to $L_{\text{Context}_i}$, we also apply the standard loss on the training sub-example in \cref{loss_train}.

\begin{equation}
    \begin{aligned}
        L_{\text{Train}_i} = \text{CrossEntropy}(\hat{t}_{\text{O}_{i}}^{(N+1)}, t_{\text{O}_{i}}^{(N+1)}) \\
    \end{aligned}
    \label{loss_train}
\end{equation}

Lastly, we sum both losses to create the final loss $L_i=L_{\text{Context}_i} + L_{\text{Train}_i}$, where $L_{\text{Context}_i}$ can be thought of as a regularization for the standard loss $L_{\text{Train}_i}$.

As explained in \cref{sec:our_method:input}, each training example $X_{\text{Train}_i}$ contains $N+1$ sub-labels, from $N$ sub-examples in the context and one training sub-example. 
However, not all sub-examples should be weighted equally. 
For instance, the last sub-example is more important as it is located in the location of the test examples. 
Additionally, sub-examples closer to the end of the context carry more importance \citep{zhao2021calibrate}.
Thus, we apply exponential loss weight decay starting from the end of the context and decaying towards the beginning, while keeping $L_{\text{Train}_i}$ unchanged.
Formally, each sub-example $k$ is multiplied by $\gamma^j$, where $j=N+1-k$.
For example, the last sub-example is multiplied by $\gamma^1$, and the second-to-last by $\gamma^2$, and so on.
The decay is shown in \cref{loss_context} as $\omega_k$.

\subsubsection{Controlled Token Embedding Optimization}
\label{sec:our_method:updates}
As mentioned in \cref{sec:our_method:loss}, we use all the labels in each $X_{\text{Train}_i}$ to optimize the tokens within the context. 
However, some tokens in the context represent labels, and we do not update these label tokens, as they carry valuable ground truth information.
Instead, we update the other tokens in the context, carefully managing these updates to ensure controlled and precise modifications.

The controlled modification is designed with two key objectives.
First, we trust the user to provide meaningful examples representing the task, so the context should stay close to the user's intent, minimizing significant changes.
Second, few-shot optimization can lead to overfitting without proper regularization.
Controlled modification addresses both issues: it acts as a regularization mechanism while preventing overfitting.
For instance, as changes become smaller, our method converges to ICL, which is robust against overfitting.
We achieve this by using projected gradient descent, which limits each token's embedding change to an $\ell_2$ norm of size $\epsilon$ after each optimization step.

\section{Experimental Setup}
\label{sec:experimental_setup}
In this section, we provide details regarding the datasets, models, baselines, and evaluation used in our experiments. Implementation details are provided in appendix G.

\subsection{Datasets}
\label{sec:experimental_setup:datasets}
In this work, we focus on a classification task and select a variety of datasets to ensure robust conclusions across different task types. 
We include \sst \citep{socher-etal-2013-recursive} for sentiment analysis, \agnews \citep{Zhang2015CharacterlevelCN} for news classification, \dbpedia \citep{NIPS2015_250cf8b5} for ontology classification, and \trec \citep{li-roth-2002-learning} for question classification. 
These datasets represent a diverse range of natural language classification tasks, include different number of classification classes, allowing us to evaluate our method comprehensively. 
More details are provided in appendix E.

In addition, we include a new dataset that we term Set Classification. 
We created this dataset to ensure it was not part of the training process for any model, further supporting the robustness of our conclusions. 
The dataset consists of a list of random words divided into $C$ distinct sets, where each set represents a unique class labeled by another random word. All words in the sets and their corresponding class labels are unique. 
From each group, we randomly select $G$ words and assign the group label as their classification. More details are provided in appendix A.

\subsection{Models}
We use models of varying sizes and quality to ensure robust evaluation and conclusions. For the relatively small model, we use \bloomNoSpace1.7B \citep{scao2022bloom}, while for larger models, we opt for \gptNoSpace6B\citep{gpt-j-6b} and \llama 8B\citep{llama3modelcard}. The \gpt model is noted for its robust performance, while \llama is currently among the leading models in the field.

\subsection{Baselines}
We compare our method to several groups of few-shot learning techniques. In the first group, we include LoRA \citep{hu2021lora}, one of the leading efficient fine-tuning methods. Additionally, we compare against several prompt-tuning approaches, including Prompt Tuning (PT) \citep{lester2021power}, Prefix Tuning \citep{li2021prefix}, and Instruction Prompt Tuning (IPT) \citep{singhal2022large}. Finally, we compare our method to In-Context Learning (ICL) \citep{brown2020language}.

For some of the few-shot methods, we introduce an alternative version that incorporates instructions, as indicated in \cref{table:main_table} with a \textdagger. 
Instead of initializing the learnable tokens randomly, we initialize them with instructions specified in appendix D. We apply instructions to PT, IPT, and our method, reporting results for both random and instruction-based prompt initialization.

\begin{table}[t!]
\begin{center}
\vspace{\upTabAboveCap}
\caption{
    \textbf{Baseline Comparisons} Mean accuracy of various methods and our \AlgoNameNoSpace, across several models and datasets. Evaluations are conducted using 2, 4, and 6 shots.
}
\label{table:main_table}
\vspace{\upTabBelowCap}
\resizebox{\columnwidth}{!}{%

\begin{tabular}{llccccccccc}

\hline\noalign{\smallskip}\hline

\rowcolor{gray!5}  & & \multicolumn{3}{|c|}{\bloom 1.7B} & \multicolumn{3}{|c|}{\gpt 6B} & \multicolumn{3}{|c}{\llama 8B}\\

\rowcolor{gray!5}  \multirow{-2}{*}{Dataset} & \multirow{-2}{*}{Method} & \multicolumn{1}{|c}{2} & 4 & \multicolumn{1}{c|}{6} & \multicolumn{1}{|c}{2} & 4 & \multicolumn{1}{c|}{6} & \multicolumn{1}{|c}{2} & 4 & \multicolumn{1}{c}{6} \\

\hline\noalign{\smallskip}\hline

\multirow{9}{*}{\sstNoSpace} & Prefix & $47.80$ & $47.33$ & $49.00$ & $52.23$ & $52.50$ & $52.87$ & $-$ & $-$ & $-$ \\

& ICL & $50.53$ & $60.83$ & $61.87$ & $50.57$ & $67.47$ & $77.47$ & $76.43$ & $80.63$ & $83.10$ \\

 & PT\textdagger & $64.97$ & $65.07$ & $65.07$ & $57.10$ & $52.93$ & $55.70$ & $72.97$ & $73.47$ & $84.57$ \\

 & PT & $56.03$ & $56.90$ & $58.33$ & $64.07$ & $64.37$ & $64.60$ & $64.27$ & $65.70$ & $67.03$ \\

 & IPT\textdagger & $58.50$ & $61.83$ & $62.80$ & $51.50$ & $\textbf{83.20}$ & $84.80$ & $86.90$ & $88.03$ & $94.40$ \\

 & IPT & $48.50$ & $58.80$ & $61.87$ & $48.13$ & $82.27$ & $87.17$ & $57.20$ & $87.40$ & $90.43$ \\
 
 & LoRA & $\textbf{66.40}$ & $66.93$ & $66.90$ & $\textbf{69.80}$ & $71.53$ & $73.17$ & $68.73$ & $71.27$ & $83.97$ \\

 \rowcolor{golden!10} \cellcolor{white} &  CPT\textdagger & $59.53$ & $\textbf{72.40}$ & $\textbf{74.83}$ & $52.53$ & $82.03$ & $\textbf{88.07}$ & $\textbf{92.73}$ & $95.07$ & $96.40$ \\

 \rowcolor{golden!10} \cellcolor{white} & CPT & $50.77$ & $70.70$ & $74.10$ & $50.53$ & $82.90$ & $88.03$ & $83.83$ & $\textbf{96.30}$ & $\textbf{96.50}$ \\

\cline{2-11}\noalign{\smallskip}

\multirow{9}{*}{\agnewsNoSpace} & Prefix & $24.87$ & $25.35$ & $26.02$ & $32.32$ & $33.33$ & $46.08$ & $-$ & $-$ & $-$  \\
 
 & ICL & $35.12$ & $34.28$ & $42.48$ & $66.73$ & $62.38$ & $69.57$ & $79.38$ & $82.32$ & $85.27$ \\

 & PT\textdagger & $28.67$ & $30.73$ & $41.17$ & $37.85$ & $44.85$ & $62.92$ & $59.60$ & $57.02$ & $68.02$ \\

 & PT & $33.57$ & $36.98$ & $\textbf{56.08}$ & $56.85$ & $56.13$ & $75.10$ & $69.32$ & $67.92$ & $69.33$ \\

 & IPT\textdagger & $36.95$ & $31.90$ & $42.93$ & $67.02$ & $63.00$ & $74.85$ & $82.93$ & $\textbf{84.45}$ & $85.08$ \\

 & IPT & $38.77$ & $38.20$ & $47.78$ & $66.02$ & $63.92$ & $74.00$ & $80.52$ & $76.30$ & $80.98$ \\
 
 & LoRA & $29.50$ & $30.80$ & $33.98$ & $56.12$ & $56.03$ & $72.55$ & $70.62$ & $74.97$ & $73.70$ \\

 \rowcolor{golden!10} \cellcolor{white} & CPT\textdagger & $33.68$ & $33.13$ & $41.10$ & $71.35$ & $\textbf{68.73}$ & $75.68$ & $83.17$ & $84.28$ & $84.67$ \\

 \rowcolor{golden!10} \cellcolor{white} & CPT & $\textbf{40.85}$ & $\textbf{44.48}$ & $50.40$ & $\textbf{74.80}$ & $68.62$ & $\textbf{76.22}$ & $\textbf{83.78}$ & $81.92$ & $\textbf{85.43}$ \\

\cline{2-11}\noalign{\smallskip}

\multirow{9}{*}{\dbpediaNoSpace} & Prefix & $19.76$ & $19.74$ & $23.65$ & $13.25$ & $16.43$ & $24.94$ & $-$ & $-$ & $-$ \\
  
& ICL & $48.20$ & $51.40$ & $55.17$ & $50.87$ & $62.46$ & $70.76$ & $71.66$ & $72.44$ & $79.93$ \\

 & PT\textdagger & $24.90$ & $26.32$ & $34.75$ & $21.01$ & $22.12$ & $37.44$ & $55.30$ & $57.21$ & $66.26$ \\

 & PT & $46.71$ & $41.94$ & $45.93$ & $23.39$ & $29.69$ & $40.53$ & $55.81$ & $52.72$ & $55.02$ \\

 & IPT\textdagger & $33.28$ & $40.36$ & $45.85$ & $47.10$ & $67.60$ & $75.09$ & $81.10$ & $87.69$ & $92.06$ \\

 & IPT & $48.09$ & $54.60$ & $70.57$ & $52.86$ & $67.27$ & $70.73$ & $72.92$ & $76.11$ & $78.44$ \\
 
 & LoRA & $43.30$ & $41.13$ & $41.18$ & $30.15$ & $28.02$ & $41.50$ & $54.24$ & $59.50$ & $63.21$ \\

\rowcolor{golden!10} \cellcolor{white} & CPT\textdagger & $33.80$ & $48.13$ & $51.18$ & $53.20$ & $\textbf{77.30}$ & $\textbf{81.00}$ & $\textbf{84.23}$ & $\textbf{90.33}$ & $\textbf{93.08}$ \\

\rowcolor{golden!10} \cellcolor{white} & CPT & $\textbf{58.85}$ & $\textbf{65.78}$ & $\textbf{73.55}$ & $\textbf{68.29}$ & $75.07$ & $77.65$ & $77.38$ & $78.49$ & $82.42$ \\

\cline{2-11}\noalign{\smallskip}

\multirow{9}{*}{\trecNoSpace} & Prefix & $19.10$ & $24.49$ & $29.92$ & $30.76$ & $30.04$ & $27.87$ & $-$ & $-$ & $-$ \\

 & ICL & $33.54$ & $33.33$ & $28.53$ & $28.94$ & $35.14$ & $32.49$ & $35.32$ & $42.48$ & $40.34$ \\

 & PT\textdagger & $30.91$ & $33.70$ & $39.31$ & $29.02$ & $34.66$ & $43.89$ & $43.42$ & $48.81$ & $51.73$ \\

 & PT & $32.18$ & $32.26$ & $35.69$ & $31.16$ & $32.79$ & $37.86$ & $32.77$ & $33.98$ & $33.83$ \\

 & IPT\textdagger & $27.83$ & $36.64$ & $42.92$ & $31.04$ & $43.12$ & $43.09$ & $51.72$ & $62.14$ & $65.13$ \\

 & IPT & $32.37$ & $36.59$ & $42.60$ & $29.59$ & $38.90$ & $40.38$ & $36.94$ & $45.62$ & $52.08$ \\

 & LoRA & $34.07$ & $33.22$ & $33.50$ & $34.17$ & $33.73$ & $37.63$ & $31.21$ & $33.21$ & $36.36$ \\

\rowcolor{golden!10} \cellcolor{white} & CPT\textdagger & $29.72$ & $35.64$ & $\textbf{45.38}$ & $33.39$ & $44.20$ & $\textbf{45.83}$ & $\textbf{57.26}$ & $\textbf{67.00}$ & $\textbf{69.29}$ \\

\rowcolor{golden!10} \cellcolor{white} & CPT & $\textbf{35.68}$ & $\textbf{41.79}$ & $45.16$ & $\textbf{35.37}$ & $\textbf{44.66}$ & $42.71$ & $45.12$ & $57.54$ & $60.18$ \\

\hline\noalign{\smallskip}\hline\noalign{\smallskip}

\end{tabular}%
}

\vspace{\upTabBelowTab}

\end{center}
\end{table}

\subsection{Evaluation}
We evaluate each model and dataset using three different numbers of training samples: 2, 4, and 6. For each configuration, the reported results are averaged accuracy over 30 experiments, consisting of 10 randomly sampled templates and 3 different random seeds, with the templates described in appendix F. By utilizing randomized seeds, we ensure variation in the selection of training examples. This extensive setup is crucial for achieving a comprehensive and robust evaluation, especially given that these methods are known to be highly sensitive to the selection of training examples and templates \citep{voronov2024mind, zhao2021calibrate}. Further evaluation details can be found in appendix C.

\section{Results}
\label{sec:results}


\subsection{Main Results}
\label{exp:results:main_results}

In \cref{table:main_table}, we demonstrate that \AlgoName convincingly performs better than the baselines in most cases, with particularly pronounced gains in harder tasks. Furthermore, \AlgoName's performance becomes more efficient and effective as the models grow stronger, such as with \llamaNoSpace.

\textbf{Better With Harder Tasks} \AlgoName demonstrates improvements across various datasets, with more pronounced gains in tasks we define as harder based on two factors: the number of shots and the number of classes. As illustrated in \cref{table:main_table}, task difficulty increases with fewer shots and more classes. For example, on the \dbpedia dataset, which has 14 classes, decreasing the shots from 6 to 4 widens the performance gap between \AlgoName and the baselines from (3, 6, 1) to (11, 10, 3) across the models: \bloomNoSpace, \gptNoSpace, and \llamaNoSpace.

\textbf{Decisive Advantage with Powerful Models} The strength of the model plays a significant role in performance. 
As the model becomes better, \AlgoNameNoSpace's advantage becomes more pronounced across all datasets and shot settings. 
For instance, \llama consistently outperforms other baselines across all datasets, except in one case where results are comparable.
With \gptNoSpace, a slightly older model, the results are lower in two instances, with one comparable outcome, both on \sst, the easier task as previously discussed. 
When comparing with \bloom, the weakest model in our comparison, we observe lower performance on two occasions, specifically on the two easier datasets.







\begin{figure*}[t!]
    \centering
    \vspace{\upFigureAboveFig}
    \includegraphics[width=\textwidth]{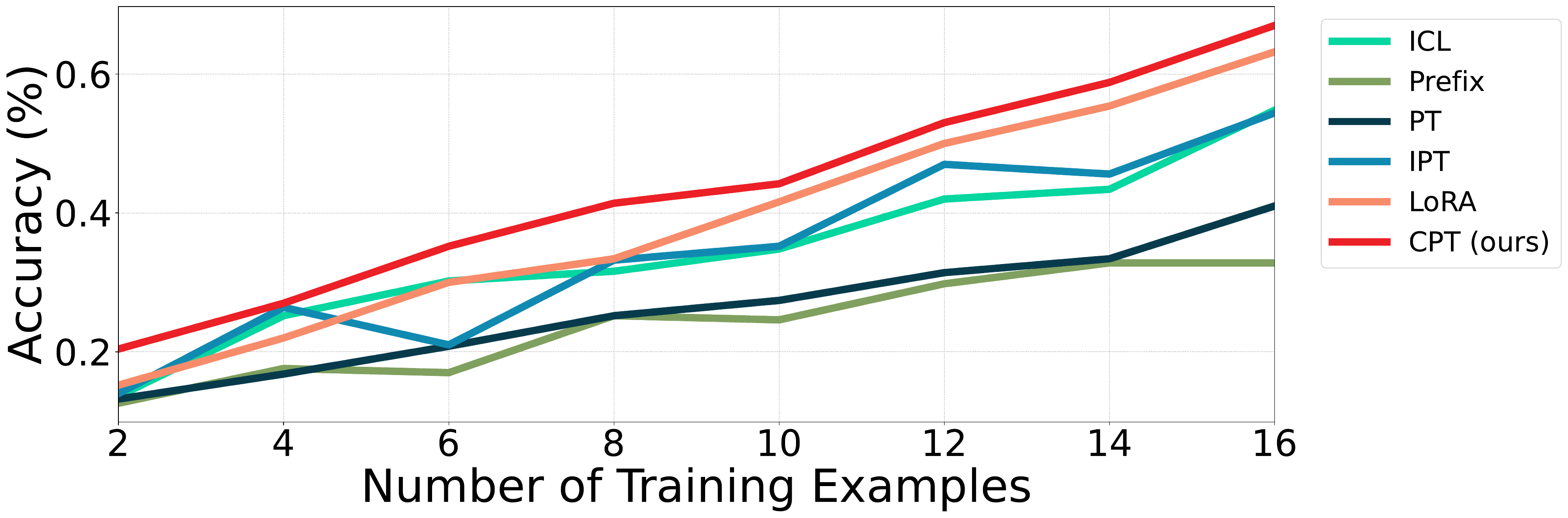}
    \vspace{\upFigureBelowFig}
    \caption{\textbf{Set Classification Dataset} Mean accuracy of our \AlgoName method versus baselines as the number of training examples increases using the \gpt model.}
    \vspace{\upFigureBelowCap}
    \label{fig:eval_language}
\end{figure*}

\begin{figure*}[b!]
    \centering
    \vspace{\downFigureAboveFig}
    \includegraphics[width=\textwidth]{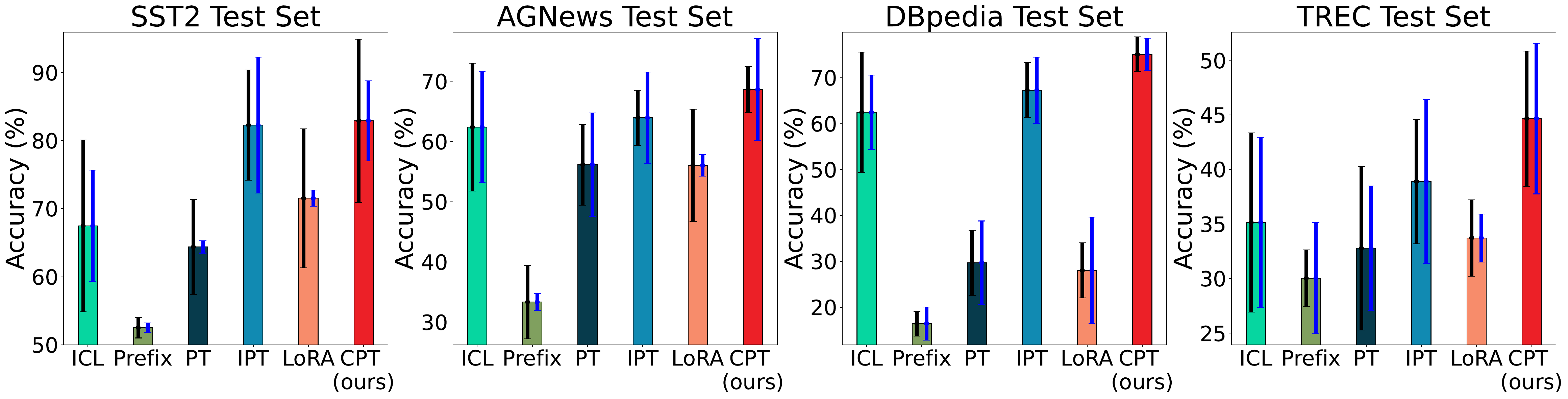}
    \vspace{\downFigureBelowFig}
    \caption{\textbf{Accuracy and Standard Deviation} Comparison of accuracy and standard deviation between \AlgoName and baselines, evaluated with 4-shot on \gpt model. The black bars represent the mean std across different templates, while the blue bars represent the mean std across different seeds. }
    \vspace{\downFigureBelowCap}
    \label{fig:eval_std}
\end{figure*}

\subsection{Set Classification}
\label{exp:results:set_classification}

We use the Set Classification dataset, a new classification dataset introduced in \cref{sec:experimental_setup:datasets}, to verify that \AlgoNameNoSpace's improvements are not due to prior knowledge encoded in the model from training data. 
This ensures that the observed performance gains are truly a result of our method's effectiveness and not influenced by pre-existing biases or memorized information.

In \cref{fig:eval_language}, we compare the mean accuracy of \AlgoName to the baselines across different numbers of training examples. 
As shown, \AlgoName consistently outperforms the baselines, even in this entirely novel dataset, further validating its robustness and generalization capabilities across unseen tasks.

\subsection{Standard Deviation}
\label{exp:results:standard_deviation}

Standard deviation (std) plays a crucial role in few-shot learning due to the sensitivity of these methods to both the training examples and the chosen template \citep{zhao2021calibrate, voronov2024mind}. In \cref{fig:eval_std}, we present accuracy along with two types of std bars: black bars represent the mean std across different templates, while blue bars represent the mean std across different seeds. We demonstrate that \AlgoName significantly improves accuracy across various models and datasets in a statistically significant manner. More information is presented in appendix B.

Our method's standard deviation performs equivalently to other methods in most cases, while in certain cases, such as with \dbpediaNoSpace, \AlgoName exhibits both higher accuracy and lower std, reinforcing its robustness in complex tasks. 
However, the sensitivity of our method does not follow a clear pattern across random seeds or templates. 
For instance, while randomness in templates and training examples has an equal influence on std in \dbpedia and \trec, \sst shows a higher std for template randomness, and \agnews is more sensitive to variations in training examples.

\subsection{Ablations}
\label{sec:results:ablations}



The most important design choices that positively impacted \AlgoNameNoSpace's performance are the loss design and the projections. 
These improvements are evident across 2, 4, and 6-shot settings, as shown in \cref{table:ablation}.
Different options for the loss design are specified under ``Loss Tokens'', with three configurations: using only the training label, using the training label plus one random context label, and using the training label plus all context labels. The latter significantly outperforms the training-only configuration.
The ablation over the projection is specified under ``Input $\epsilon$'' and ``Format $\epsilon$'' demonstrating that both too small a change (which converges to ICL) and too large a norm are suboptimal.
Lastly, we examined the effect of ``Loss Weighting''.
We propose three options: Mean, which applies uniform weighting across all labels; Equal, which assigns equal weight to both the training label loss and the context label losses; and the option we use, Decay, which reduces the influence of context labels further from the training example. 
On this dataset, Decay works slightly better, and in most cases, the improvement is more significant.

In addition to these core design choices, we explored several alternative configurations that ultimately did not enhance performance. 
Under ``Loss Weighting'', we experimented with the "Equal" option, which assigns equal loss weight to both the training example loss and the entire context loss, where the training loss can be multiplied by the noted value (\emph{e.g.}, 1, 10). 
We also tested the projection type ``All-Tokens'' which applies the projection to the entire context collectively rather than token-by-token. 
Under ``Updated Tokens'' we attempted to modify only specific parts of the context. 
Additionally, under ``Mask Training'' we masked the training example from the context to prevent the model from simply copying the answer. 
However, none of these configurations led to performance improvements.

\begin{table}[t!]
    \centering
    \vspace{\upTabAboveCap}
    \caption{\textbf{Ablation Study} We present the mean accuracy for various ablations using the \gpt model and the \dbpedia dataset, including loss tokens (train example, random, or all context), loss weighting (decay and mean), projection type (token-wise or all-tokens), epsilon values for input and format, updated tokens (input, format, masks), and masking of the training example.}
    \vspace{\upTabBelowCap}
    
    \resizebox{1.0\columnwidth}{!}{%
    \begin{tabular}{llllllc>{\centering\arraybackslash}p{0.083\textwidth}>{\centering\arraybackslash}p{0.083\textwidth}>{\centering\arraybackslash}p{0.081\textwidth}}

        \hline\noalign{\smallskip}\hline
        \rowcolor{gray!5}  & & & & & & & \multicolumn{3}{c}{\makecell{Number of Training Examples}} \\
        \rowcolor{gray!5} \multirow{-2}{*}{Loss Tokens}  & \multirow{-2}{*}{Loss Weighting} & \multirow{-2}{*}{Projection Type} & \multirow{-2}{*}{Input $\epsilon$} & \multirow{-2}{*}{Format $\epsilon$} &  \multirow{-2}{*}{Updated Tokens} & \multirow{-2}{*}{Mask Training Example} & $2$ & $4$ & $6$ \\

        \hline\noalign{\smallskip}\hline

        Train Example & \multirow{3}{*}{Decay $0.95$} & \multirow{3}{*}{Token-Wise} & \multirow{3}{*}{0.1} & \multirow{3}{*}{0.1} & \multirow{3}{*}{Input \& Format} & \multirow{3}{*}{\xmark} & $58.09$ & $61.54$ & $66.69$ \\
        Train Example \& 1 Random & & & & & & & $69.48$ & $72.08$ & $76.80$ \\
        Train Example \& All Context & & & & & & & $69.54$ & $73.03$ & $76.58$ \\
        
        \hline

        \multirow{6}{*}{Train Example \& All Context}
         & Mean & \multirow{6}{*}{Token-Wise} & \multirow{6}{*}{0.1} & \multirow{6}{*}{0.1} & \multirow{6}{*}{Input \& Format} & \multirow{6}{*}{\xmark} & $69.62$ & $72.91$ & $76.49$ \\
        
         & Equal $1$ & & & & & & $69.07$ & $72.82$ & $76.23$ \\
         & Equal $10$ & & & & & & $69.35$ & $71.01$ & $75.11$ \\
        
         & Decay $0.99$ & & & & & & $69.59$ & $72.97$ & $76.43$ \\
         & Decay $0.95$ & & & & & & $69.54$ & $73.03$ & $76.58$ \\
         & Decay $0.5$ & & & & & & $69.60$ & $72.39$ & $76.44$ \\

        \hline

        \multirow{4}{*}{Train Example \& All Context}
         & \multirow{4}{*}{Decay 0.95} & \multirow{4}{*}{All-Tokens} & 0.001 & - & \multirow{4}{*}{Input \& Format} & \multirow{4}{*}{\xmark} & $51.52$ & $63.41$ & $71.50$ \\
         & & & 0.01 & - & & & $56.37$ & $68.12$ & $73.66$ \\
         & & & 0.1 & - & & & $69.51$ & $72.64$ & $76.06$ \\
         & & & 1.0 & - & & & $63.11$ & $64.78$ & $71.94$ \\

        \hline
        
         \multirow{5}{*}{Train Example \& All Context} & \multirow{5}{*}{Decay $0.95$} & \multirow{5}{*}{Token-Wise} & $0.01$ & $0.1$ & \multirow{5}{*}{Input \& Format} & \multirow{6}{*}{\xmark} & $65.61$ & $70.12$ & $75.63$ \\
         & & & $0.1$ & $0.1$ & & & $69.54$ & $73.03$ & $76.58$ \\
         & & & $1.0$ & $0.1$ & & & $65.29$ & $66.30$ & $73.63$ \\  
         & & & $0.1$ & $0.01$ & & & $69.53$ & $73.55$ & $76.55$ \\
         & & & $0.1$ & $1.0$ & & & $68.27$ & $71.91$ & $68.27$ \\

        \hline
        
         \multirow{3}{*}{Train Example \& All Context} & \multirow{3}{*}{Decay $0.95$} & \multirow{3}{*}{Token-Wise} & \multirow{3}{*}{0.1} & \multirow{3}{*}{0.1} & Input & \multirow{3}{*}{\xmark} & $69.47$ & $74.13$ & $76.63$ \\
         & & & & & Masks & & $63.74$ & $69.21$ & $74.91$ \\
         & & & & & Input \& Format & & $69.54$ & $73.03$ & $76.58$ \\

        \hline
        
         \multirow{1}{*}{Train Example \& All Context} & \multirow{1}{*}{Decay $0.95$} & \multirow{1}{*}{Token-Wise} & \multirow{1}{*}{0.1} & \multirow{1}{*}{0.1} & Input \& Format & \multirow{1}{*}{\cmark} & $67.55$ & $64.26$ & $68.58$ \\

        \hline

    \end{tabular}
    }   
    \vspace{\upTabBelowTab}
    \label{table:ablation}
\end{table}







\section{Discussions}
\label{sec:discussions}

\AlgoName demonstrates significant advancements in few-shot learning by integrating ICL with PT and adversarial strategies, refining context embeddings. 
Unlike traditional fine-tuning and other parameter-efficient approaches, \AlgoName optimizes only the context tokens, making it particularly effective in few-shot settings, where overfitting is a concern. \AlgoName achieves improved generalization across a wide variety of tasks, demonstrates a significant advancement, and offers meaningful insights for future few-shot learning methods.

\paragraph{Limitation \& Future Work} The computational cost associated with the iterative optimization of context embeddings is significant compared to ICL. Additionally, similar to ICL and IPT, \AlgoName is limited in the number of examples it can handle, as memory consumption scales with context length. In contrast, traditional methods are better suited for larger datasets. Future work could explore more efficient optimization strategies to reduce computational overhead and improve scalability.


\bibliography{iclr2024_conference}

\begin{thebibliography}{50}
\providecommand{\natexlab}[1]{#1}
\providecommand{\url}[1]{\texttt{#1}}
\expandafter\ifx\csname urlstyle\endcsname\relax
  \providecommand{\doi}[1]{doi: #1}\else
  \providecommand{\doi}{doi: \begingroup \urlstyle{rm}\Url}\fi

\bibitem[AI@Meta(2024)]{llama3modelcard}
AI@Meta.
\newblock Llama 3 model card.
\newblock 2024.
\newblock URL \url{https://github.com/meta-llama/llama3/blob/main/MODEL_CARD.md}.

\bibitem[Athalye et~al.(2018)Athalye, Engstrom, Ilyas, and Kwok]{athalye2018synthesizing}
Anish Athalye, Logan Engstrom, Andrew Ilyas, and Kevin Kwok.
\newblock Synthesizing robust adversarial examples.
\newblock In \emph{International conference on machine learning}, pp.\  284--293. PMLR, 2018.

\bibitem[Biggio et~al.(2013)Biggio, Corona, Maiorca, Nelson, {\v{S}}rndi{\'c}, Laskov, Giacinto, and Roli]{biggio2013evasion}
Battista Biggio, Igino Corona, Davide Maiorca, Blaine Nelson, Nedim {\v{S}}rndi{\'c}, Pavel Laskov, Giorgio Giacinto, and Fabio Roli.
\newblock Evasion attacks against machine learning at test time.
\newblock In \emph{Joint European conference on machine learning and knowledge discovery in databases}, pp.\  387--402. Springer, 2013.

\bibitem[Blau et~al.(2022)Blau, Ganz, Kawar, Bronstein, and Elad]{blau2022threat}
Tsachi Blau, Roy Ganz, Bahjat Kawar, Alex Bronstein, and Michael Elad.
\newblock Threat model-agnostic adversarial defense using diffusion models.
\newblock \emph{arXiv preprint arXiv:2207.08089}, 2022.

\bibitem[Blau et~al.(2023)Blau, Ganz, Baskin, Elad, and Bronstein]{blau2023classifier}
Tsachi Blau, Roy Ganz, Chaim Baskin, Michael Elad, and Alex Bronstein.
\newblock Classifier robustness enhancement via test-time transformation.
\newblock \emph{arXiv preprint arXiv:2303.15409}, 2023.

\bibitem[Brown et~al.(2020)Brown, Mann, Ryder, Subbiah, Kaplan, Dhariwal, Neelakantan, Shyam, Sastry, Askell, et~al.]{brown2020language}
Tom Brown, Benjamin Mann, Nick Ryder, Melanie Subbiah, Jared~D Kaplan, Prafulla Dhariwal, Arvind Neelakantan, Pranav Shyam, Girish Sastry, Amanda Askell, et~al.
\newblock Language models are few-shot learners.
\newblock \emph{Advances in neural information processing systems}, 33:\penalty0 1877--1901, 2020.

\bibitem[Carlini \& Wagner(2017)Carlini and Wagner]{carlini2017adversarial}
Nicholas Carlini and David Wagner.
\newblock Adversarial examples are not easily detected: Bypassing ten detection methods.
\newblock In \emph{Proceedings of the 10th ACM workshop on artificial intelligence and security}, pp.\  3--14, 2017.

\bibitem[Deng et~al.(2022)Deng, Wang, Hsieh, Wang, Guo, Shu, Song, Xing, and Hu]{deng2022rlprompt}
Mingkai Deng, Jianyu Wang, Cheng-Ping Hsieh, Yihan Wang, Han Guo, Tianmin Shu, Meng Song, Eric~P Xing, and Zhiting Hu.
\newblock Rlprompt: Optimizing discrete text prompts with reinforcement learning.
\newblock \emph{arXiv preprint arXiv:2205.12548}, 2022.

\bibitem[Ding et~al.(2022)Ding, Qin, Yang, Wei, Yang, Su, Hu, Chen, Chan, Chen, et~al.]{ding2022delta}
Ning Ding, Yujia Qin, Guang Yang, Fuchao Wei, Zonghan Yang, Yusheng Su, Shengding Hu, Yulin Chen, Chi-Min Chan, Weize Chen, et~al.
\newblock Delta tuning: A comprehensive study of parameter efficient methods for pre-trained language models.
\newblock \emph{arXiv preprint arXiv:2203.06904}, 2022.

\bibitem[Ghosal et~al.(2024)Ghosal, Basu, Feizi, and Manocha]{ghosal2024intcoop}
Soumya~Suvra Ghosal, Samyadeep Basu, Soheil Feizi, and Dinesh Manocha.
\newblock Intcoop: Interpretability-aware vision-language prompt tuning.
\newblock \emph{arXiv preprint arXiv:2406.13683}, 2024.

\bibitem[Goodfellow et~al.(2014)Goodfellow, Shlens, and Szegedy]{goodfellow2014explaining}
Ian~J Goodfellow, Jonathon Shlens, and Christian Szegedy.
\newblock Explaining and harnessing adversarial examples.
\newblock \emph{arXiv preprint arXiv:1412.6572}, 2014.

\bibitem[Gowal et~al.(2020)Gowal, Qin, Uesato, Mann, and Kohli]{gowal2020uncovering}
Sven Gowal, Chongli Qin, Jonathan Uesato, Timothy Mann, and Pushmeet Kohli.
\newblock Uncovering the limits of adversarial training against norm-bounded adversarial examples.
\newblock \emph{arXiv preprint arXiv:2010.03593}, 2020.

\bibitem[Gu et~al.(2021)Gu, Han, Liu, and Huang]{gu2021ppt}
Yuxian Gu, Xu~Han, Zhiyuan Liu, and Minlie Huang.
\newblock Ppt: Pre-trained prompt tuning for few-shot learning.
\newblock \emph{arXiv preprint arXiv:2109.04332}, 2021.

\bibitem[Han et~al.(2022)Han, Zhao, Ding, Liu, and Sun]{han2022ptr}
Xu~Han, Weilin Zhao, Ning Ding, Zhiyuan Liu, and Maosong Sun.
\newblock Ptr: Prompt tuning with rules for text classification.
\newblock \emph{AI Open}, 3:\penalty0 182--192, 2022.

\bibitem[Houlsby et~al.(2019)Houlsby, Giurgiu, Jastrzebski, Morrone, De~Laroussilhe, Gesmundo, Attariyan, and Gelly]{houlsby2019parameter}
Neil Houlsby, Andrei Giurgiu, Stanislaw Jastrzebski, Bruna Morrone, Quentin De~Laroussilhe, Andrea Gesmundo, Mona Attariyan, and Sylvain Gelly.
\newblock Parameter-efficient transfer learning for nlp.
\newblock In \emph{International conference on machine learning}, pp.\  2790--2799. PMLR, 2019.

\bibitem[Howard \& Ruder(2018)Howard and Ruder]{howard2018universal}
Jeremy Howard and Sebastian Ruder.
\newblock Universal language model fine-tuning for text classification.
\newblock \emph{arXiv preprint arXiv:1801.06146}, 2018.

\bibitem[Hu et~al.(2021)Hu, Shen, Wallis, Allen-Zhu, Li, Wang, Wang, and Chen]{hu2021lora}
Edward~J. Hu, Yelong Shen, Phillip Wallis, Zeyuan Allen-Zhu, Yuanzhi Li, Shean Wang, Lu~Wang, and Weizhu Chen.
\newblock Lora: Low-rank adaptation of large language models, 2021.

\bibitem[Karimi~Mahabadi et~al.(2021)Karimi~Mahabadi, Henderson, and Ruder]{karimi2021compacter}
Rabeeh Karimi~Mahabadi, James Henderson, and Sebastian Ruder.
\newblock Compacter: Efficient low-rank hypercomplex adapter layers.
\newblock \emph{Advances in Neural Information Processing Systems}, 34:\penalty0 1022--1035, 2021.

\bibitem[Khashabi et~al.(2021)Khashabi, Lyu, Min, Qin, Richardson, Welleck, Hajishirzi, Khot, Sabharwal, Singh, et~al.]{khashabi2021prompt}
Daniel Khashabi, Shane Lyu, Sewon Min, Lianhui Qin, Kyle Richardson, Sean Welleck, Hannaneh Hajishirzi, Tushar Khot, Ashish Sabharwal, Sameer Singh, et~al.
\newblock Prompt waywardness: The curious case of discretized interpretation of continuous prompts.
\newblock \emph{arXiv preprint arXiv:2112.08348}, 2021.

\bibitem[Kopiczko et~al.(2023)Kopiczko, Blankevoort, and Asano]{kopiczko2023vera}
Dawid~J. Kopiczko, Tijmen Blankevoort, and Yuki~M. Asano.
\newblock Vera: Vector-based random matrix adaptation, 2023.

\bibitem[Kurakin et~al.(2016)Kurakin, Goodfellow, and Bengio]{kurakin2016adversarial}
Alexey Kurakin, Ian Goodfellow, and Samy Bengio.
\newblock Adversarial machine learning at scale.
\newblock \emph{arXiv preprint arXiv:1611.01236}, 2016.

\bibitem[Lan et~al.(2019)Lan, Chen, Goodman, Gimpel, Sharma, and Soricut]{lan2019albert}
Zhenzhong Lan, Mingda Chen, Sebastian Goodman, Kevin Gimpel, Piyush Sharma, and Radu Soricut.
\newblock Albert: A lite bert for self-supervised learning of language representations.
\newblock \emph{arXiv preprint arXiv:1909.11942}, 2019.

\bibitem[Lester et~al.(2021)Lester, Al-Rfou, and Constant]{lester2021power}
Brian Lester, Rami Al-Rfou, and Noah Constant.
\newblock The power of scale for parameter-efficient prompt tuning, 2021.

\bibitem[Li \& Liang(2021)Li and Liang]{li2021prefix}
Xiang~Lisa Li and Percy Liang.
\newblock Prefix-tuning: Optimizing continuous prompts for generation.
\newblock \emph{arXiv preprint arXiv:2101.00190}, 2021.

\bibitem[Li \& Roth(2002)Li and Roth]{li-roth-2002-learning}
Xin Li and Dan Roth.
\newblock Learning question classifiers.
\newblock In \emph{{COLING} 2002: The 19th International Conference on Computational Linguistics}, 2002.
\newblock URL \url{https://www.aclweb.org/anthology/C02-1150}.

\bibitem[Liu et~al.(2022)Liu, Tam, Muqeeth, Mohta, Huang, Bansal, and Raffel]{liu2022few}
Haokun Liu, Derek Tam, Mohammed Muqeeth, Jay Mohta, Tenghao Huang, Mohit Bansal, and Colin~A Raffel.
\newblock Few-shot parameter-efficient fine-tuning is better and cheaper than in-context learning.
\newblock \emph{Advances in Neural Information Processing Systems}, 35:\penalty0 1950--1965, 2022.

\bibitem[Liu et~al.(2023)Liu, Zheng, Du, Ding, Qian, Yang, and Tang]{liu2023gpt}
Xiao Liu, Yanan Zheng, Zhengxiao Du, Ming Ding, Yujie Qian, Zhilin Yang, and Jie Tang.
\newblock Gpt understands, too.
\newblock \emph{AI Open}, 2023.

\bibitem[Liu et~al.(2019)Liu, Ott, Goyal, Du, Joshi, Chen, Levy, Lewis, Zettlemoyer, and Stoyanov]{liu2019roberta}
Yinhan Liu, Myle Ott, Naman Goyal, Jingfei Du, Mandar Joshi, Danqi Chen, Omer Levy, Mike Lewis, Luke Zettlemoyer, and Veselin Stoyanov.
\newblock Roberta: A robustly optimized bert pretraining approach.
\newblock \emph{arXiv preprint arXiv:1907.11692}, 2019.

\bibitem[Madry et~al.(2017)Madry, Makelov, Schmidt, Tsipras, and Vladu]{madry2017towards}
Aleksander Madry, Aleksandar Makelov, Ludwig Schmidt, Dimitris Tsipras, and Adrian Vladu.
\newblock Towards deep learning models resistant to adversarial attacks.
\newblock \emph{arXiv preprint arXiv:1706.06083}, 2017.

\bibitem[Nguyen et~al.(2015)Nguyen, Yosinski, and Clune]{nguyen2015deep}
Anh Nguyen, Jason Yosinski, and Jeff Clune.
\newblock Deep neural networks are easily fooled: High confidence predictions for unrecognizable images.
\newblock In \emph{Proceedings of the IEEE conference on computer vision and pattern recognition}, pp.\  427--436, 2015.

\bibitem[OpenAI(2024)]{openai2024chatgpt}
OpenAI.
\newblock Chatgpt (september 19 version).
\newblock \url{https://chat.openai.com}, 2024.
\newblock Large language model.

\bibitem[Peng et~al.(2023)Peng, Wang, Chen, Li, Qi, Wang, Wu, Zeng, Xu, Hou, et~al.]{peng2023does}
Hao Peng, Xiaozhi Wang, Jianhui Chen, Weikai Li, Yunjia Qi, Zimu Wang, Zhili Wu, Kaisheng Zeng, Bin Xu, Lei Hou, et~al.
\newblock When does in-context learning fall short and why? a study on specification-heavy tasks.
\newblock \emph{arXiv preprint arXiv:2311.08993}, 2023.

\bibitem[Perez et~al.(2021)Perez, Kiela, and Cho]{perez2021true}
Ethan Perez, Douwe Kiela, and Kyunghyun Cho.
\newblock True few-shot learning with language models.
\newblock \emph{Advances in neural information processing systems}, 34:\penalty0 11054--11070, 2021.

\bibitem[Radford et~al.(2019)Radford, Wu, Child, Luan, Amodei, Sutskever, et~al.]{radford2019language}
Alec Radford, Jeffrey Wu, Rewon Child, David Luan, Dario Amodei, Ilya Sutskever, et~al.
\newblock Language models are unsupervised multitask learners.
\newblock \emph{OpenAI blog}, 1\penalty0 (8):\penalty0 9, 2019.

\bibitem[Raffel et~al.(2020)Raffel, Shazeer, Roberts, Lee, Narang, Matena, Zhou, Li, and Liu]{raffel2020exploring}
Colin Raffel, Noam Shazeer, Adam Roberts, Katherine Lee, Sharan Narang, Michael Matena, Yanqi Zhou, Wei Li, and Peter~J Liu.
\newblock Exploring the limits of transfer learning with a unified text-to-text transformer.
\newblock \emph{Journal of machine learning research}, 21\penalty0 (140):\penalty0 1--67, 2020.

\bibitem[Rebuffi et~al.(2021)Rebuffi, Gowal, Calian, Stimberg, Wiles, and Mann]{rebuffi2021fixing}
Sylvestre-Alvise Rebuffi, Sven Gowal, Dan~A Calian, Florian Stimberg, Olivia Wiles, and Timothy Mann.
\newblock Fixing data augmentation to improve adversarial robustness.
\newblock \emph{arXiv preprint arXiv:2103.01946}, 2021.

\bibitem[Scao et~al.(2022)Scao, Fan, Akiki, and et~al.]{scao2022bloom}
Teven~Le Scao, Angela Fan, Christopher Akiki, and et~al.
\newblock Bloom: A 176b-parameter open-access multilingual language model.
\newblock \emph{arXiv preprint arXiv:2211.05100}, 2022.

\bibitem[Singhal et~al.(2022)Singhal, Azizi, Tu, Mahdavi, Wei, Chung, Scales, Tanwani, Cole-Lewis, Pfohl, et~al.]{singhal2022large}
Karan Singhal, Shekoofeh Azizi, Tao Tu, S~Sara Mahdavi, Jason Wei, Hyung~Won Chung, Nathan Scales, Ajay Tanwani, Heather Cole-Lewis, Stephen Pfohl, et~al.
\newblock Large language models encode clinical knowledge.
\newblock \emph{arXiv preprint arXiv:2212.13138}, 2022.

\bibitem[Socher et~al.(2013)Socher, Perelygin, Wu, Chuang, Manning, Ng, and Potts]{socher-etal-2013-recursive}
Richard Socher, Alex Perelygin, Jean Wu, Jason Chuang, Christopher~D. Manning, Andrew Ng, and Christopher Potts.
\newblock Recursive deep models for semantic compositionality over a sentiment treebank.
\newblock In \emph{Proceedings of the 2013 Conference on Empirical Methods in Natural Language Processing}, pp.\  1631--1642, Seattle, Washington, USA, October 2013. Association for Computational Linguistics.
\newblock URL \url{https://www.aclweb.org/anthology/D13-1170}.

\bibitem[Sun et~al.(2019)Sun, Qiu, Xu, and Huang]{sun2019fine}
Chi Sun, Xipeng Qiu, Yige Xu, and Xuanjing Huang.
\newblock How to fine-tune bert for text classification?
\newblock In \emph{Chinese computational linguistics: 18th China national conference, CCL 2019, Kunming, China, October 18--20, 2019, proceedings 18}, pp.\  194--206. Springer, 2019.

\bibitem[Sun et~al.(2023)Sun, Liu, Iter, Zhu, and Iyyer]{sun2023does}
Simeng Sun, Yang Liu, Dan Iter, Chenguang Zhu, and Mohit Iyyer.
\newblock How does in-context learning help prompt tuning?
\newblock \emph{arXiv preprint arXiv:2302.11521}, 2023.

\bibitem[Voronov et~al.(2024)Voronov, Wolf, and Ryabinin]{voronov2024mind}
Anton Voronov, Lena Wolf, and Max Ryabinin.
\newblock Mind your format: Towards consistent evaluation of in-context learning improvements.
\newblock \emph{arXiv preprint arXiv:2401.06766}, 2024.

\bibitem[Wang \& Komatsuzaki(2021)Wang and Komatsuzaki]{gpt-j-6b}
Ben Wang and Aran Komatsuzaki.
\newblock Gpt-j-6b: A 6 billion parameter autoregressive language model.
\newblock \url{https://github.com/kingoflolz/mesh-transformer-jax}, 2021.
\newblock Accessed: 2024-05-26.

\bibitem[Wang et~al.(2023)Wang, Panda, Karlinsky, Feris, Sun, and Kim]{wang2023multitask}
Zhen Wang, Rameswar Panda, Leonid Karlinsky, Rogerio Feris, Huan Sun, and Yoon Kim.
\newblock Multitask prompt tuning enables parameter-efficient transfer learning.
\newblock \emph{arXiv preprint arXiv:2303.02861}, 2023.

\bibitem[Wang \& Liang(2024)Wang and Liang]{wang2024lora}
Zhengbo Wang and Jian Liang.
\newblock Lora-pro: Are low-rank adapters properly optimized?
\newblock \emph{arXiv preprint arXiv:2407.18242}, 2024.

\bibitem[Zaken et~al.(2021)Zaken, Ravfogel, and Goldberg]{zaken2021bitfit}
Elad~Ben Zaken, Shauli Ravfogel, and Yoav Goldberg.
\newblock Bitfit: Simple parameter-efficient fine-tuning for transformer-based masked language-models.
\newblock \emph{arXiv preprint arXiv:2106.10199}, 2021.

\bibitem[Zhang et~al.(2022)Zhang, Roller, Goyal, Artetxe, Chen, Chen, Dewan, Diab, Li, Lin, et~al.]{zhang2022opt}
Susan Zhang, Stephen Roller, Naman Goyal, Mikel Artetxe, Moya Chen, Shuohui Chen, Christopher Dewan, Mona Diab, Xian Li, Xi~Victoria Lin, et~al.
\newblock Opt: Open pre-trained transformer language models.
\newblock \emph{arXiv preprint arXiv:2205.01068}, 2022.

\bibitem[Zhang et~al.(2015{\natexlab{a}})Zhang, Zhao, and LeCun]{NIPS2015_250cf8b5}
Xiang Zhang, Junbo Zhao, and Yann LeCun.
\newblock Character-level convolutional networks for text classification.
\newblock In C.~Cortes, N.~Lawrence, D.~Lee, M.~Sugiyama, and R.~Garnett (eds.), \emph{Advances in Neural Information Processing Systems}, volume~28. Curran Associates, Inc., 2015{\natexlab{a}}.
\newblock URL \url{https://proceedings.neurips.cc/paper_files/paper/2015/file/250cf8b51c773f3f8dc8b4be867a9a02-Paper.pdf}.

\bibitem[Zhang et~al.(2015{\natexlab{b}})Zhang, Zhao, and LeCun]{Zhang2015CharacterlevelCN}
Xiang Zhang, Junbo~Jake Zhao, and Yann LeCun.
\newblock Character-level convolutional networks for text classification.
\newblock In \emph{NIPS}, 2015{\natexlab{b}}.

\bibitem[Zhao et~al.(2021)Zhao, Wallace, Feng, Klein, and Singh]{zhao2021calibrate}
Zihao Zhao, Eric Wallace, Shi Feng, Dan Klein, and Sameer Singh.
\newblock Calibrate before use: Improving few-shot performance of language models.
\newblock In \emph{International conference on machine learning}, pp.\  12697--12706. PMLR, 2021.

\end{thebibliography}
\bibliographystyle{iclr2024_conference}


\end{document}


\maketitle

\appendix

\section{Set Classification}
\label{app:set_classification}
The Set Classification dataset was specifically designed to ensure that no prior model had been exposed to the data during the training process, thereby validating that our method remains effective in such cases. This ensures that the classification process is based purely on the patterns learned during the few-shot learning tasks. The dataset is composed of two groups of words: one group consists of the class words, which serve as the labels, and the other group consists of the class member words, where all the words are distinct.

For constructing the dataset, each group of words is controlled by three key numbers provided by the user: the total number of groups $C$, the number of words per group $N$, and the number of words in each input sequence $M$. For example, in the experiments conducted in Fig. 5, we used $C=10$ classes (groups), where each class group contained $N=5$ distinct words, and each input sequence fed into the model was composed of $M=4$ randomly selected words from a group, and the dataset presented in \cref{app:used_dataset}. This design ensures a clear separation between classes and provides a robust setting to evaluate the model’s ability to classify novel sets of words based on limited exposure.

\begin{table}[h!]
\centering
\caption{Class Labels and Corresponding Class Members}
\begin{tabular}{|c|p{10cm}|}
\hline
\textbf{Class} & \textbf{Class Members} \\ 
\hline
apple & pillow, lantern, teleseme, crayon, monkey \\
\hline
tornado & Jacobinism, lupoid, elevator, Octogynia, zoo \\
\hline
window & surfboard, poststertorous, deforciant, dicatalexis, pirate \\
\hline
pizza & indevout, cookie, helmet, balneotherapeutics, opusculum \\
\hline
moon & Pygopodidae, hydramnios, circumnutate, spydom, punnigram \\
\hline
bat & histology, door, howitzer, lighthouse, carmoisin \\
\hline
iceberg & triketo, trepidatory, dulcifluous, knotwort, house \\
\hline
flower & airmark, violin, nerval, patter, key \\
\hline
castle & Swietenia, nonliability, unbrotherliness, pharyngotonsillitis, robot \\
\hline
lion & whisky, orange, spoon, antivaccinist, television \\
\hline
\end{tabular}

\label{app:used_dataset}
\end{table}

\clearpage
\section{Standard Deviation}
\label{app:standard_deviation}

\begin{table}[ht!]
\begin{center}
\caption{
    \textbf{Standard Deviation Analysis}
    Standard deviations (STD) corresponding to Table 1. Each experiment shows three STD values separated by a backslash: (1) STD over 30 experiments with 10 random templates and 3 seeds, (2) mean STD over templates, and (3) mean STD over seeds.
}

\resizebox{\columnwidth}{!}{%

\begin{tabular}{llccccccccc}
\hline\noalign{\scriptstyleskip}\hline

\rowcolor{gray!5}  & & \multicolumn{9}{c}{Model} \\
\rowcolor{gray!5}  & & \multicolumn{3}{|c|}{\bloom 1.7B} & \multicolumn{3}{|c|}{\gpt 6B} & \multicolumn{3}{|c}{\llama 8B}\\
\rowcolor{gray!5}  \multirow{-3}{*}{Dataset} & \multirow{-3}{*}{Method} & 2 & 4 & 6 & 2 & 4 & 6 & 2 & 4 & 6 \\

\hline\noalign{\scriptstyleskip}\hline\noalign{\scriptstyleskip}

\multirow{9}{*}{\sstNoSpace} & Prefix Tuning & $00.5/00.4/00.1$ & $03.1/02.7/01.9$ & $03.1/02.6/02.1$ & $00.8/00.6/00.2$ & $02.3/01.5/00.7$ & $05.8/04.3/03.9$ & $-$ & $-$ & $-$ \\

& ICL & $04.3/04.0/01.5$ & $12.6/08.6/09.4$ & $14.9/10.6/09.7$ & $05.5/04.0/03.0$ & $14.1/12.6/08.2$ & $13.1/09.9/09.9$ & $13.2/12.7/06.1$ & $15.7/11.9/10.2$ & $13.7/12.2/06.5$ \\

 & PT\textdagger & $07.6/07.6/00.4$ & $08.2/08.2/00.7$ & $08.1/08.1/00.6$ & $06.8/06.5/02.3$ & $07.8/06.9/04.3$ & $09.5/09.0/04.9$ & $16.6/16.5/01.0$ & $17.1/17.1/01.6$ & $12.7/10.9/05.7$ \\

 & PT & $08.6/08.5/01.7$ & $08.9/08.6/02.4$ & $08.7/08.4/02.5$ & $07.7/07.6/01.3$ & $07.0/07.0/00.9$ & $07.4/07.4/01.0$ & $06.4/06.1/03.1$ & $06.8/06.6/02.8$ & $07.0/06.5/03.7$ \\

 & IPT\textdagger & $12.3/12.3/03.4$ & $14.1/11.4/09.8$ & $15.3/10.1/12.6$ & $05.8/05.1/02.5$ & $11.1/08.0/08.3$ & $12.3/12.3/02.0$ & $08.7/06.8/05.6$ & $12.5/11.2/07.3$ & $02.7/02.6/01.6$ \\

 & IPT & $02.0/01.5/00.4$ & $11.6/08.7/08.2$ & $14.8/09.0/11.8$ & $00.7/00.4/00.2$ & $13.0/08.1/10.0$ & $07.2/05.7/04.7$ & $13.7/13.7/03.9$ & $10.6/09.6/05.5$ & $12.0/10.3/05.3$ \\
 
 & LoRA & $06.9/06.9/00.2$ & $06.5/06.5/00.5$ & $06.4/06.3/00.5$ & $09.5/09.5/00.8$ & $10.2/10.2/01.2$ & $10.0/09.9/01.4$ & $11.4/11.4/01.4$ & $15.0/14.6/07.6$ & $12.1/11.9/07.2$ \\

 & \rowcolor{golden!10} CPT\textdagger & $12.3/12.3/03.6$ & $13.8/09.8/10.5$ & $15.4/14.2/09.6$ & $07.6/06.7/03.2$ & $11.1/09.1/07.0$ & $07.0/06.2/04.1$ & $05.0/04.3/02.5$ & $04.1/03.0/02.3$ & $02.0/01.7/01.2$ \\

 & \rowcolor{golden!10} CPT & $07.7/05.3/02.9$ & $12.0/11.1/06.9$ & $12.9/10.5/09.7$ & $05.5/03.9/02.9$ & $12.7/12.0/05.9$ & $10.7/08.9/05.1$ & $13.2/10.9/08.5$ & $01.6/01.5/01.0$ & $01.3/01.2/01.0$ \\

\cline{2-11}\noalign{\scriptstyleskip}

\multirow{9}{*}{\agnewsNoSpace} & Prefix Tuning & $01.7/01.7/00.6$ & $05.8/02.9/05.2$ & $05.3/03.9/03.9$ & $05.9/05.9/00.5$ & $06.2/06.1/01.4$ & $12.7/09.4/08.2$ &  $-$ & $-$ & $-$ \\

& ICL & $10.5/06.8/08.7$ & $11.7/10.1/06.4$ & $12.2/11.1/06.0$ & $10.0/08.9/05.2$ & $13.3/10.6/09.2$ & $10.4/09.4/05.0$ & $08.8/03.2/08.1$ & $03.2/03.0/02.3$ & $03.1/02.7/02.4$ \\

 & PT\textdagger & $05.2/04.1/04.1$ & $06.0/04.3/04.4$ & $10.9/10.5/07.3$ & $16.2/16.1/00.9$ & $13.7/11.5/09.5$ & $13.6/11.9/07.8$ & $11.9/11.8/06.1$ & $10.9/10.3/05.8$ & $10.5/09.0/06.5$ \\

 & PT & $07.9/04.6/06.9$ & $08.4/06.8/05.7$ & $09.9/09.3/05.2$ & $12.3/11.5/05.3$ & $10.8/06.7/08.6$ & $07.0/02.5/06.8$ & $15.0/15.0/01.5$ & $15.4/15.2/02.7$ & $12.4/12.3/02.2$ \\

 & IPT\textdagger & $11.7/09.2/08.4$ & $07.5/05.5/05.0$ & $15.0/09.4/12.4$ & $11.0/08.3/07.9$ & $07.1/03.1/06.5$ & $07.6/03.5/06.8$ & $02.7/02.5/01.6$ & $03.5/02.8/02.2$ & $03.4/02.9/02.5$ \\

 & IPT & $12.0/08.6/08.9$ & $10.6/09.1/07.2$ & $12.0/09.4/07.9$ & $11.1/10.0/06.3$ & $08.8/04.6/07.6$ & $07.4/04.6/05.7$ & $03.6/03.1/01.8$ & $08.1/03.2/07.6$ & $05.2/02.2/04.7$ \\
 
 & LoRA & $03.2/03.2/00.5$ & $06.8/03.3/06.1$ & $04.6/04.4/02.7$ & $09.0/09.0/01.3$ & $09.4/09.3/01.8$ & $08.8/05.3/07.3$ & $15.0/15.0/01.5$ & $15.4/15.2/02.7$ & $12.4/12.3/02.2$ \\

 & \rowcolor{golden!10} CPT\textdagger & $07.2/05.7/05.4$ & $06.0/03.4/04.8$ & $11.9/08.5/09.2$ & $09.6/07.2/07.1$ & $09.0/03.6/08.5$ & $09.2/04.4/08.1$ & $03.1/02.5/02.2$ & $02.9/02.6/02.4$ & $03.4/01.8/03.2$ \\

 & \rowcolor{golden!10} CPT & $12.8/08.7/10.3$ & $12.2/07.8/10.3$ & $11.3/08.9/07.1$ & $08.6/05.4/07.4$ & $09.1/03.8/08.5$ & $07.2/03.7/06.2$ & $03.3/02.6/02.3$ & $04.3/03.6/02.8$ & $02.9/02.3/02.3$ \\

\cline{2-11}\noalign{\scriptstyleskip}

\multirow{9}{*}{\dbpediaNoSpace} & Prefix Tuning & $03.5/03.4/01.9$ & $06.4/03.1/05.7$ & $08.7/03.6/08.1$ & $02.3/02.3/01.7$ & $04.5/02.7/03.6$ & $07.9/04.9/06.2$ & $-$ & $-$ & $-$ \\

& ICL & $24.1/23.3/06.9$ & $25.8/23.5/08.9$ & $23.9/23.6/06.0$ & $16.6/16.3/05.9$ & $15.7/13.1/08.1$ & $06.7/05.8/04.0$ & $07.7/06.4/06.2$ & $06.8/02.6/06.5$ & $04.2/02.3/04.0$ \\

 & PT\textdagger & $15.6/08.5/13.2$ & $09.7/09.6/01.7$ & $07.1/04.6/05.8$ & $10.3/10.3/00.9$ & $06.3/05.8/04.2$ & $06.3/05.8/04.2$ & $19.5/17.3/11.0$ & $15.7/12.7/09.0$ & $15.4/13.7/08.2$ \\

 & PT & $11.2/11.1/04.2$ & $10.8/10.8/01.3$ & $13.0/12.4/05.9$ & $09.9/08.2/06.4$ & $11.3/07.1/09.1$ & $08.9/05.0/07.5$ & $11.9/11.7/04.2$ & $15.7/15.3/02.9$ & $13.5/13.4/01.5$ \\

 & IPT\textdagger & $25.6/21.2/14.7$ & $24.3/22.6/08.9$ & $27.3/26.2/08.5$ & $16.4/15.5/05.7$ & $11.0/09.7/06.2$ & $06.8/05.3/05.2$ & $05.3/04.3/03.0$ & $04.5/03.8/02.7$ & $04.5/04.1/01.9$ \\

 & IPT & $26.2/25.1/07.5$ & $25.0/20.3/11.9$ & $07.6/07.0/02.9$ & $12.2/11.2/06.0$ & $09.6/06.0/07.2$ & $05.4/03.7/04.1$ & $09.7/08.4/04.6$ & $06.0/03.6/05.1$ & $05.6/03.1/05.3$ \\
 
 & LoRA & $11.4/11.0/03.1$ & $11.6/11.6/00.3$ & $11.7/11.7/00.4$ & $11.6/10.9/04.9$ & $13.0/06.0/11.6$ & $09.8/06.0/07.9$ & $13.1/13.0/01.7$ & $14.3/14.2/02.2$ & $13.7/13.7/01.5$ \\

 & \rowcolor{golden!10} CPT\textdagger & $23.2/14.5/18.0$ & $12.0/10.1/07.1$ & $22.1/20.1/10.5$ & $15.6/08.6/14.3$ & $06.5/05.0/04.5$ & $06.0/03.2/05.2$ & $06.2/05.6/02.7$ & $03.8/03.4/02.4$ & $02.3/02.2/01.7$ \\

 & \rowcolor{golden!10} CPT & $15.5/13.4/06.5$ & $11.8/08.8/07.1$ & $04.7/04.0/02.5$ & $10.9/08.2/05.5$ & $05.0/03.8/03.5$ & $03.9/03.0/03.0$ & $06.1/05.0/04.8$ & $04.3/03.5/03.0$ & $04.3/02.6/03.5$ \\

\cline{2-11}\noalign{\scriptstyleskip}

\multirow{9}{*}{\trecNoSpace} & Prefix Tuning & $06.7/00.8/06.6$ & $06.4/02.9/06.0$ & $07.0/04.6/06.0$ & $03.1/02.0/02.5$ & $05.9/02.6/05.1$ & $03.7/03.6/00.8$ & $-$ & $-$ & $-$ \\

& ICL & $11.0/07.2/08.2$ & $10.5/08.2/06.8$ & $13.8/09.0/09.1$ & $08.9/05.9/06.8$ & $11.0/08.2/07.8$ & $12.6/08.3/09.4$ & $08.6/05.6/06.3$ & $14.2/07.9/12.0$ & $13.2/08.3/10.7$ \\

 & PT\textdagger & $05.9/04.4/03.7$ & $07.5/06.7/04.5$ & $11.2/08.7/07.8$ & $05.5/04.3/03.4$ & $06.2/06.0/04.3$ & $13.5/08.2/11.7$ & $09.5/05.7/08.3$ & $11.3/06.1/09.4$ & $10.3/08.4/08.1$ \\

 & PT & $03.8/03.4/01.5$ & $08.2/07.3/06.7$ & $11.2/08.6/09.1$ & $04.0/04.0/00.9$ & $08.1/07.5/05.7$ & $09.7/08.2/07.8$ & $05.0/05.0/01.5$ & $04.5/04.5/02.5$ & $03.8/03.8/02.0$ \\

 & IPT\textdagger & $05.5/03.6/04.0$ & $10.1/09.1/07.1$ & $16.8/08.1/15.5$ & $06.8/05.3/04.2$ & $07.7/05.1/05.9$ & $14.0/07.8/11.9$ & $13.6/06.3/12.3$ & $09.7/05.9/08.6$ & $07.2/05.4/05.5$ \\

 & IPT & $10.5/07.1/07.8$ & $06.1/05.9/05.0$ & $14.1/07.4/12.8$ & $09.7/05.4/08.0$ & $09.3/05.7/07.5$ & $13.0/03.8/12.5$ & $12.1/08.5/07.8$ & $11.5/09.1/08.0$ & $14.0/05.4/13.3$ \\
 
 & LoRA & $03.9/03.9/01.0$ & $04.0/04.0/00.3$ & $04.1/04.1/00.4$ & $02.5/02.5/00.4$ & $03.6/03.5/02.2$ & $11.9/07.2/10.4$ & $03.3/03.3/01.0$ & $03.5/02.7/02.5$ & $16.5/08.1/15.4$ \\

 & \rowcolor{golden!10} CPT\textdagger & $08.0/05.8/05.4$ & $07.7/06.9/06.3$ & $09.9/07.0/07.9$ & $08.5/05.7/06.4$ & $12.9/08.3/10.6$ & $11.2/08.2/09.0$ & $13.1/06.9/11.6$ & $09.8/03.7/09.2$ & $05.0/04.1/03.4$ \\

 & \rowcolor{golden!10} CPT & $09.1/05.2/07.3$ & $07.9/07.2/05.6$ & $12.9/07.0/10.8$ & $07.4/04.1/06.1$ & $08.7/06.2/06.9$ & $08.6/05.5/07.3$ & $16.8/08.4/14.5$ & $07.4/06.5/05.8$ & $07.9/05.6/06.5$ \\

\hline\noalign{\scriptstyleskip}\hline\noalign{\scriptstyleskip}

\end{tabular}%
}
\label{table:main_table_std}
\end{center}
\end{table}

In \cref{table:main_table_std} we present the standard deviations (STD) corresponding to the main results shown in Table 1. For each experiment, we display three STD values, separated by a backslash. These values represent the variability in the results across different configurations:

1. The first value shows the standard deviation over 30 experiments, which includes 10 random templates and 3 seeds that determine the training examples.
2. The second value provides the mean of the standard deviation over the templates, the standard deviation across 10 templates, and the mean of the standard deviation across 3 seeds.
3. The third value presents the mean standard deviation over the seeds, the standard deviation over 3 seeds, and the mean over 10 templates.

This detailed breakdown of standard deviations allows for a more thorough understanding of the variability in model performance across different templates and seeds.

\clearpage
\section{Evaluation Details}
\label{app:evaluation_details}
All the graphs and ablation studies were conducted and evaluated using the DBPedia dataset with the GPT-J model. This setup was chosen due to the diversity of the DBPedia dataset, which includes a broad range of categories and entities, making it an ideal candidate for comprehensive evaluation. The use of GPT-J, a powerful generative model, ensures that the results are reflective of state-of-the-art performance in language modeling tasks. The combination of DBPedia and GPT-J allows us to thoroughly investigate the behavior of the model across various ablation settings, ensuring robust insights into the performance of different methods and configurations.

\subsection{Pruning for Classification}

In our evaluation setup, we use pruning for classification by focusing only on the first token of the label, which is unique across all datasets. A common approach in the in-context learning setup is to iterate over all possible labels for each test sample and select the label with the highest probability according to the language model (LM). However, this approach can become computationally expensive, especially in cases where there are a large number of classes.

Similarly to \cite{ratner2022parallel}, and given that the first token in each dataset is unique, we predict only the first token of the label and perform classification based on this value. While this approach deviates slightly from the common practice of iterating over all possible labels, the effect on the results should be minor.

\subsection{Test Set Size}

For our experiments, we used a varying number of test examples depending on the dataset. Specifically, we used 100 test examples for the SST-2 dataset, and for datasets with a larger number of classes, the number of test examples was scaled linearly with the number of classes. For example, in the DBpedia dataset, which has 7 times more classes than SST-2, we used 700 test examples to ensure that the evaluation is proportional to the number of classes. This scaling helps to maintain a balanced evaluation across datasets with differing complexities, ensuring robust performance metrics for each method.

\clearpage
\section{Instruction Details}
\label{app:instruction_details}

In some of the experiments, we use specific instructions to guide the model in performing the classification tasks. Below in \cref{table:inst} that shows the instructions used for each dataset across all relevant methods:

\begin{table}[h]
\centering
\caption{Instructions used for relevant datasets in the experiments.}
\begin{tabular}{|c|p{12cm}|}
\hline
\textbf{Dataset} & \textbf{Instruction} \\
\hline
SST2 & Classify the sentiment of the following text as positive or negative: \\
\hline
AG News & Classify the following text into one of the following categories: World, Sports, Business, Technology \\
\hline
DBpedia & Classify the following text into one of the following categories: Company, Educational Institution, Artist, Athlete, Office Holder, Mean Of Transportation, Building, Natural Place, Village, Animal, Plant, Album, Film, Written Work \\
\hline
TREC & Classify the following text into one of the following categories: Description, Entity, Expression, Human, Location, Number \\

\hline
\end{tabular}
\label{table:inst}
\end{table}

\clearpage
\section{Dataset Details}
\label{app:dataset_details}
\begin{table}[t!]
\centering
\caption{
\textbf{Dataset Overview} These are the datasets used, representing a range of different types of classification tasks, including \sstNoSpace, \agnewsNoSpace, \dbpediaNoSpace, and \trecNoSpace. Each dataset has a varying number of classes (denoted by $|C|$).
}

\resizebox{0.5\columnwidth}{!}{%
\begin{tabular}{ccc}

\hline\noalign{\smallskip}\hline
\rowcolor{gray!5} Dataset & Task & $|C|$  \\ 

\hline\noalign{\smallskip}\hline

\sst & Sentiment analysis (movie) & 2 \\

\agnews & News classification (topic) & 4 \\

\dbpedia & Ontology classification & 14 \\

\trec & Question classification (answer type) & 6 \\

\hline
\end{tabular}
}

\label{table:datasets}
\end{table}

In our experiments, we used four different datasets, each representing a unique classification task. Table \ref{table:datasets} provides an overview of the datasets and their respective tasks. Each dataset has a varying number of classes, denoted by \(|C|\), which are detailed below:

\begin{itemize}
    \item \textbf{SST-2}: This dataset is used for \textit{sentiment analysis}, where the task is to classify movie reviews as either positive or negative. It contains 2 distinct classes. \vspace{0.2cm}
    \item \textbf{AG News}: The AG News dataset is used for \textit{news classification}. The task is to classify news articles into one of four categories: World, Sports, Business, and Technology. This dataset contains 4 classes. \vspace{0.2cm}
    \item \textbf{DBpedia}: The DBpedia dataset is focused on \textit{ontology classification}. The task involves classifying textual content into one of 14 distinct categories, which include entities such as Company, Artist, Village, and more. \vspace{0.2cm}
    \item \textbf{TREC}: This dataset is used for \textit{question classification}, where the goal is to classify questions into one of 6 answer types, including Description, Entity, Human, and Location. \vspace{0.2cm}
\end{itemize}

Each dataset contains a specific number of examples based on its classification task, allowing us to evaluate the model’s performance across a diverse range of challenges.

\clearpage
\section{Template Details}
\label{app:example_templates}

\begin{table}[ht!]
\centering
\caption{\textbf{Template Options for Various Datasets} We provide various template options for different datasets. Each dataset include both input and output templates, and also includes intra-separators between inputs and labels, as well as inter-separators between examples.}

\resizebox{\columnwidth}{!}{%
\begin{tabular}{ccccc}

\hline\noalign{\smallskip}\hline
\rowcolor{gray!5} Dataset & Input Template & Intra-Separator & Output Template & Inter-Separator \\ 

\hline\noalign{\smallskip}\hline

\multirow{1}{*}{\sst} & \multirow{4}{*}{\makecell{"input: \{\}", \\ "text: {}", \\ "sentence: \{\}", \\ "\{\}" }} & \multirow{4}{*}{\makecell{" ", \\ "$\backslash$n"}} & \makecell{"output: \{\}", "target: \{\}", "label: \{\}", \\ "emotion: \{\}", "semtiment: \{\}", "A \{\} one.", \\ "It was \{\}.", "All in all \{\}.", "A \{\} piece."} & \multirow{4}{*}{\makecell{" ", \\ "$\backslash$n", \\ "$\backslash$n$\backslash$n"}}\\

\cline{1-1}\noalign{\vskip 1pt} \cline{4-4}\noalign{\vskip 1pt}

\multirow{1}{*}{\agnews} & & & \multirow{3}{*}{\makecell{"output: \{\}", "target: \{\}", "label: \{\}", \\ "Topic: \{\}.", "Subject: \{\}.", \\ "This is about \{\}.", "It is about \{\}."}} \\

\cline{1-1}\noalign{\vskip 1pt}

\multirow{1}{*}{\dbpedia} & \\

\cline{1-1}\noalign{\vskip 1pt}

\multirow{1}{*}{\trec} & \\

\hline
\end{tabular}
}

\label{table:templates}
\end{table}

In our experiments, we use randomly selected templates from the options provided in \cref{table:templates}, suggested in \cite{voronov2024mind}. Each dataset is associated with both input and output templates, which are used to format the input data and the expected output during few-shot learning tasks.

\begin{itemize}
    \item \textbf{Input Template}: As shown, this column lists the different templates for formatting the input data. For example, the SST-2 dataset uses "input: {}" and "text: {}" as input templates to introduce the input text.
    \item \textbf{Intra-Separator}: This separator is used between components (input and output) within a single example. For instance, AG News uses "\textbackslash n" as an intra-separator between the input sentence and the output label.
    \item \textbf{Output Template}: The output template defines how the expected output is structured. For example, SST-2 employs formats like "output: {}, target: {}, label: {}" to guide the model in generating structured output.
    \item \textbf{Inter-Separator}: This column represents the separator used between multiple examples during training. In datasets like AG News and DBpedia, "\textbackslash n\textbackslash n" is used to separate examples.
\end{itemize}

We randomly select templates from the ones listed in \cref{table:templates} for each experiment. This randomness in selecting templates introduces variability in the prompts, making the evaluation more robust and testing the model’s ability to generalize across different input-output structures.

\clearpage
\section{Implementation Details}
\label{app:implementation_details}

\begin{table}[b!]
\begin{center}
\caption{
    \textbf{Hyperparameters} Hyperparameters used for each experiment across 2, 4, and 6 shots for different models, including \bloom 1.7B, \gpt 6B, and \llama 8B. The table shows learning rates (lr), epsilon values for input and format, and other parameters for methods such as Prefix Tuning, Prompt Tuning, IPT, LoRA, and CPT. The experiments were conducted on datasets like SST-2, AG News, DBpedia, and TREC.
    
}
\resizebox{\columnwidth}{!}{%

\begin{tabular}{lllccccccccc}
\hline\noalign{\scriptstyleskip}\hline

\rowcolor{gray!5}  & & & \multicolumn{3}{|c|}{\bloom 1.7B} & \multicolumn{3}{|c|}{\gpt 6B} & \multicolumn{3}{|c}{\llama 8B}\\
\rowcolor{gray!5}  \multirow{-2}{*}{Dataset} & \multirow{-2}{*}{Method} & \multirow{-2}{*}{Paremeter} & \multicolumn{1}{|c}{2} & 4 & \multicolumn{1}{c|}{6} & \multicolumn{1}{|c}{2} & 4 & \multicolumn{1}{c|}{6} & \multicolumn{1}{|c}{2} & 4 & \multicolumn{1}{c}{6} \\

\hline\noalign{\smallskip}\hline

\multirow{12}{*}{\sstNoSpace} & Prefix Tuning & lr & $1e-3$ & $1e-3$ & $1e-3$ & $1e-5$ & $1e-4$ & $1e-3$ & $-$ & $-$ & $-$ \\

 & PT\textdagger & lr & $1e-5$ & $1e-5$ & $1e-5$ & $1e-4$ & $1e-3$ & $1e-3$ & $1e-5$ & $1e-5$ & $1e-5$ \\

 & PT & lr & $1e-5$ & $1e-5$ & $1e-5$ & $1e-5$ & $1e-5$ & $1e-5$ & $1e-5$ & $1e-5$ & $1e-5$ \\

 & IPT\textdagger & lr & $1e-5$ & $1e-4$ & $1e-4$ & $1e-5$ & $1e-3$ & $1e-4$ & $1e-5$ & $1e-5$ & $1e-4$ \\

 & IPT & lr & $1e-5$ & $1e-5$ & $1e-5$ & $1e-5$ & $1e-4$ & $1e-4$ & $1e-5$ & $1e-5$ & $1e-5$ \\
 
 & LoRA & lr & $1e-5$ & $1e-5$ & $1e-5$ & $1e-5$ & $1e-5$ & $1e-5$ & $1e-5$ & $1e-4$ & $1e-4$ \\

 & \rowcolor{golden!10} & lr & $1e-5$ & $1e-3$ & $1e-4$ & $1e-5$ & $1e-4$ & $1e-3$ & $1e-5$ & $1e-5$ & $1e-5$ \\
 
 & \rowcolor{golden!10} & Input $\epsilon$ & $1e-3$ & $1e-0$ & $1e-0$ & $1e-3$ & $1e-1$ & $1e-1$ & $1e-1$ & $1e-1$ & $1e-0$ \\

  & \multirow{-3}{*}{CPT\textdagger} \rowcolor{golden!10} & Format $\epsilon$ & $1e-3$ & $1e-3$ & $1e-3$ & $1e-3$ & $1e-2$ & $1e-3$ & $1e-2$ & $1e-1$ & $1e-0$ \\

 & \rowcolor{golden!10} & lr & $1e-3$ & $1e-3$ & $1e-4$ & $1e-5$ & $1e-4$ & $1e-4$ & $1e-3$ & $1e-4$ & $1e-4$ \\
 
 & \rowcolor{golden!10} & Input $\epsilon$ & $1e-2$ & $1e-0$ & $1e-0$ & $1e-3$ & $1e-0$ & $1e-0$ & $1e-2$ & $1e-0$ & $1e-2$ \\

  & \multirow{-3}{*}{CPT} \rowcolor{golden!10} & Format $\epsilon$ & $1e-2$ & $1e-2$ & $1e-3$ & $1e-3$ & $1e-3$ & $1e-2$ & $1e-3$ & $1e-3$ & $1e-3$ \\

\cline{2-12}\noalign{\scriptstyleskip}

\multirow{12}{*}{\agnewsNoSpace} & Prefix Tuning & lr & $1e-4$ & $1e-3$ & $1e-3$ & $1e-5$ & $1e-5$ & $1e-3$ & $-$ & $-$ & $-$ \\

 & PT\textdagger & lr & $1e-3$ & $1e-3$ & $1e-3$ & $1e-5$ & $1e-3$ & $1e-3$ & $1e-4$ & $1e-4$ & $1e-4$ \\

 & PT & lr & $1e-3$ & $1e-3$ & $1e-3$ & $1e-4$ & $1e-3$ & $1e-3$ & $1e-4$ & $1e-5$ & $1e-4$ \\

 & IPT\textdagger & lr & $1e-3$ & $1e-3$ & $1e-3$ & $1e-5$ & $1e-4$ & $1e-4$ & $1e-4$ & $1e-5$ & $1e-5$ \\

 & IPT & lr & $1e-4$ & $1e-3$ & $1e-4$ & $1e-5$ & $1e-5$ & $1e-4$ & $1e-5$ & $1e-5$ & $1e-5$ \\
 
 & LoRA & lr & $1e-5$ & $1e-4$ & $1e-3$ & $1e-5$ & $1e-5$ & $1e-4$ & $1e-5$ & $1e-5$ & $1e-5$ \\

 & \rowcolor{golden!10} & lr & $1e-4$ & $1e-3$ & $1e-3$ & $1e-4$ & $1e-4$ & $1e-4$ & $1e-5$ & $1e-5$ & $1e-5$ \\
 
 & \rowcolor{golden!10} & Input $\epsilon$ & $1e-2$ & $1e-0$ & $1e-0$ & $1e-1$ & $1e-1$ & $1e-2$ & $1e-1$ & $1e-3$ & $1e-3$ \\

  & \multirow{-3}{*}{CPT\textdagger} \rowcolor{golden!10} & Format $\epsilon$ & $1e-1$ & $1e-2$ & $1e-0$ & $1e-1$ & $1e-3$ & $1e-0$ & $1e-1$ & $1e-2$ & $1e-3$ \\

 & \rowcolor{golden!10} & lr & $1e-4$ & $1e-4$ & $1e-3$ & $1e-3$ & $1e-4$ & $1e-4$ & $1e-3$ & $1e-4$ & $1e-3$ \\
 
 & \rowcolor{golden!10} & Input $\epsilon$ & $1e-2$ & $1e-0$ & $1e-0$ & $1e-2$ & $1e-0$ & $1e-0$ & $1e-2$ & $1e-3$ & $1e-3$ \\

  & \multirow{-3}{*}{CPT} \rowcolor{golden!10} & Format $\epsilon$ & $1e-2$ & $1e-0$ & $1e-0$ & $1e-3$ & $1e-3$ & $1e-0$ & $1e-3$ & $1e-3$ & $1e-3$ \\

\cline{2-12}\noalign{\scriptstyleskip}

\multirow{12}{*}{\dbpediaNoSpace} & Prefix Tuning & lr & $1e-3$ & $1e-3$ & $1e-3$ & $1e-3$ & $1e-3$ & $1e-3$ & $-$ & $-$ & $-$ \\

 & PT\textdagger & lr & $1e-3$ & $1e-5$ & $1e-3$ & $1e-5$ & $1e-3$ & $1e-3$ & $1e-4$ & $1e-4$ & $1e-4$ \\

 & PT & lr & $1e-4$ & $1e-5$ & $1e-4$ & $1e-3$ & $1e-3$ & $1e-3$ & $1e-4$ & $1e-5$ & $1e-5$ \\

 & IPT\textdagger & lr & $1e-4$ & $1e-5$ & $1e-5$ & $1e-5$ & $1e-4$ & $1e-5$ & $1e-5$ & $1e-5$ & $1e-5$ \\

 & IPT & lr & $1e-5$ & $1e-5$ & $1e-5$ & $1e-5$ & $1e-5$ & $1e-5$ & $1e-5$ & $1e-5$ & $1e-5$ \\
 
 & LoRA & lr & $1e-4$ & $1e-5$ & $1e-5$ & $1e-4$ & $1e-4$ & $1e-4$ & $1e-5$ & $1e-5$ & $1e-5$ \\

 & \rowcolor{golden!10} & lr & $1e-5$ & $1e-5$ & $1e-5$ & $1e-4$ & $1e-5$ & $1e-5$ & $1e-5$ & $1e-5$ & $1e-5$ \\
 
 & \rowcolor{golden!10} & Input $\epsilon$ & $1e-2$ & $1e-2$ & $1e-1$ & $1e-0$ & $1e-1$ & $1e-1$ & $1e-0$ & $1e-1$ & $1e-1$ \\

  & \multirow{-3}{*}{CPT\textdagger} \rowcolor{golden!10} & Format $\epsilon$ & $1e-1$ & $1e-0$ & $1e-1$ & $1e-3$ & $1e-0$ & $1e-1$ & $1e-1$ & $1e-0$ & $1e-1$ \\

 & \rowcolor{golden!10} & lr & $1e-4$ & $1e-4$ & $1e-5$ & $1e-4$ & $1e-4$ & $1e-4$ & $1e-5$ & $1e-5$ & $1e-5$ \\
 
 & \rowcolor{golden!10} & Input $\epsilon$ & $1e-0$ & $1e-2$ & $1e-0$ & $1e-0$ & $1e-0$ & $1e-0$ & $1e-2$ & $1e-0$ & $1e-3$ \\

  & \multirow{-3}{*}{CPT} \rowcolor{golden!10} & Format $\epsilon$ & $1e-0$ & $1e-0$ & $1e-0$ & $1e-0$ & $1e-3$ & $1e-3$ & $1e-2$ & $1e-3$ & $1e-2$ \\

\cline{2-12}\noalign{\scriptstyleskip}

\multirow{12}{*}{\trecNoSpace} & Prefix Tuning & lr & $1e-3$ & $1e-3$ & $1e-3$ & $1e-3$ & $1e-3$ & $1e-5$ & $-$ & $-$ & $-$ \\

 & PT\textdagger & lr & $1e-3$ & $1e-3$ & $1e-3$ & $1e-3$ & $1e-3$ & $1e-3$ & $1e-4$ & $1e-4$ & $1e-4$ \\

 & PT & lr & $1e-5$ & $1e-3$ & $1e-3$ & $1e-5$ & $1e-3$ & $1e-3$ & $1e-5$ & $1e-5$ & $1e-5$ \\

 & IPT\textdagger & lr & $1e-3$ & $1e-3$ & $1e-3$ & $1e-4$ & $1e-3$ & $1e-4$ & $1e-4$ & $1e-4$ & $1e-5$ \\

 & IPT & lr & $1e-5$ & $1e-3$ & $1e-3$ & $1e-4$ & $1e-4$ & $1e-4$ & $1e-5$ & $1e-5$ & $1e-5$ \\
 
 & LoRA & lr & $1e-4$ & $1e-5$ & $1e-5$ & $1e-5$ & $1e-5$ & $1e-4$ & $1e-5$ & $1e-5$ & $1e-4$ \\

 & \rowcolor{golden!10} & lr & $1e-3$ & $1e-3$ & $1e-3$ & $1e-4$ & $1e-4$ & $1e-x$ & $1e-4$ & $1e-5$ & $1e-5$ \\
 
 & \rowcolor{golden!10} & Input $\epsilon$ & $1e-0$ & $1e-0$ & $1e-0$ & $1e-1$ & $1e-0$ & $1e-0$ & $1e-1$ & $1e-1$ & $1e-1$ \\

  & \multirow{-3}{*}{CPT\textdagger} \rowcolor{golden!10} & Format $\epsilon$ & $1e-3$ & $1e-1$ & $1e-2$ & $1e-1$ & $1e-0$ & $1e-2$ & $1e-3$ & $1e-0$ & $1e-0$ \\

 & \rowcolor{golden!10} & lr & $1e-3$ & $1e-3$ & $1e-4$ & $1e-3$ & $1e-3$ & $1e-3$ & $1e-4$ & $1e-4$ & $1e-4$ \\
 
 & \rowcolor{golden!10} & Input $\epsilon$ & $1e-0$ & $1e-0$ & $1e-0$ & $1e-0$ & $1e-0$ & $1e-3$ & $1e-0$ & $1e-0$ & $1e-0$ \\

  & \multirow{-3}{*}{CPT} \rowcolor{golden!10} & Format $\epsilon$ & $1e-0$ & $1e-0$ & $1e-3$ & $1e-2$ & $1e-2$ & $1e-0$ & $1e-2$ & $1e-3$ & $1e-0$ \\

\hline\noalign{\scriptstyleskip}\hline\noalign{\scriptstyleskip}

\end{tabular}%
}
\label{table:main_table_hpt}
\end{center}
\end{table}

\subsection{Hyperparameter Details}

In \cref{table:main_table_hpt} we present the hyperparameters used in our experiments across different models and datasets. The table provides the specific learning rates (`lr`), epsilon values (`$\epsilon$'), and format settings for the various methods applied to each dataset. The experiments were conducted using multiple model architectures, including \textbf{\bloom 1.7B}, \textbf{\gpt 6B}, and \textbf{\llama 8B}, and we selected the best hyperparameters for each experiment: 2, 4, and 6 shots. Below is an overview of the key hyperparameters:

\begin{itemize}
    \item \textbf{Learning Rate (`lr')}: The table provides the learning rates used for each method and dataset combination. For methods like \textit{Prefix Tuning (PT)}, \textit{Prompt Tuning (PT)}, \textit{IPT}, and \textit{LoRA}, learning rates vary from \textbf{1e-5} to \textbf{1e-3}, depending on the specific model and dataset.
    
    \item \textbf{CPT Hyperparameters}: For \textit{CPT}, we also report epsilon values (`$\epsilon$') for both the \textit{input} and the \textit{format} components. These epsilon values control the magnitude of the perturbations applied during optimization. The values of epsilon vary across different models and datasets, generally ranging from \textbf{1e-2} to \textbf{1e-0} for both input and format components.
    
    \item \textbf{Model Variability}: The table reflects variability in hyperparameter choices depending on the model size and architecture. For instance, \textit{GPT-3 6B} typically requires higher learning rates compared to \textit{BLOOM 1.7B}, as seen with \textit{CPT} and other methods. The hyperparameters are carefully tuned to optimize performance on tasks such as SST-2, AG News, DBpedia, and TREC.
\end{itemize}

These hyperparameters are critical for achieving optimal performance in few-shot learning settings. They control the learning process, model updates, and how much the model is allowed to adapt to new data. The values in \cref{table:main_table_hpt} are based on extensive experimentation and fine-tuning to ensure the best results for each method and dataset.

\subsection{Methods Implementation Details}

In our experiments, we utilized existing implementations for several methods and implemented IPT ourselves. Specifically, we used the implementations provided by the \textit{Parameter-Efficient Fine-Tuning \cite{peft} (PEFT)} library \footnote{\url{https://huggingface.co/docs/peft/en/index}} for methods such as \textbf{LoRA}, \textbf{Prefix Tuning}, and \textbf{Prompt Tuning (PT)}. For IPT, we built our implementation based on the PEFT framework.

For all experiments, we used the recommended parameters:
\begin{itemize}
    \item For LoRA, we set \(\alpha = 16\) and the rank \(r = 8\).
    \item For Prompt Tuning, Prefix Tuning, and IPT we used 8 learnable tokens.
\end{itemize}

By using the PEFT framework, we ensure that our fine-tuning processes for LoRA, Prefix Tuning, and PT are aligned with current standards, while our custom IPT implementation extends the framework to allow for additional flexibility in parameter-efficient training.

\subsection{Training Details}
We utilized the `Fine-tune a pretrained model' package from \cite{wolf-etal-2020-transformers}, which provides a comprehensive framework for training and evaluating models\footnote{\url{https://huggingface.co/docs/transformers/en/training}}. For all baselines, we employed the default parameters provided by the trainer, ensuring consistency across experiments. Each model was trained for 25 epochs, allowing sufficient time for convergence while maintaining uniform training conditions across methods.

\clearpage

\bibliography{iclr2024_conference}
\bibliographystyle{iclr2024_conference}